\documentclass[10pt,twocolumn,letterpaper]{article}

\usepackage[accsupp]{axessibility}  
\usepackage{graphicx}           
\usepackage{booktabs}           
\usepackage{multirow}           
\usepackage[table]{xcolor}  
\usepackage{multirow} 
\usepackage{pgfplots}
\usepgfplotslibrary{groupplots}
\usepackage{amssymb,pifont}
\usepackage{longtable} 
\usepackage{cvpr}      
\definecolor{cvprblue}{rgb}{0.21,0.49,0.74}
\usepackage[pagebackref,breaklinks,colorlinks,allcolors=cvprblue]{hyperref}

\def\confName{CVPR}
\def\confYear{2026}

\title{VinQA: Visual Elements Interleaved Long-form Answer Generation for Real-World Multimodal Document QA}

\author{Young Rok Jang$^{*}$ \quad Hyesoo Kong$^{*}$ \quad Kyunghwan An \quad Jae Sub Huh \\ 
Gyeonghun KIM \quad Stanley Jungkyu Choi\\
LG AI Research \\
{\tt\small \{jyrok3357, hyesoo.kong, ghkayne.kim, stanleyjk.choi\}@lgresearch.ai}
}

\begin{document}
\maketitle

\renewcommand{\thefootnote}{\fnsymbol{footnote}}
\footnotetext[1]{Equal contribution.}
\renewcommand{\thefootnote}{\arabic{footnote}}

\begin{abstract}

Real-world documents combine text with tables, charts, photographs, and diagrams arranged in diverse layouts, yet existing research on multimodal large language models (MLLMs) for document QA predominantly produces text-only responses, underutilizing these visual elements. We introduce VinQA, a dataset designed for long-form answer generation where cited visual elements are explicitly interleaved with their supporting text and grounded in relevant document pages. To support this task, we study two encoding methods for feeding raw document page images into an MLLM, along with their visual-element citation mechanisms: (1) Page Encoding, which directly encodes full-page images with bounding boxes of visual elements and treats these boxed regions as citable units; and (2) Modality Encoding, which parses each page to extract text and crop visual elements, encodes them separately, and uses these cropped elements as citable units. In our experiments, we propose M-GroSE, a multimodal evaluation framework extending GroUSE to assess such answers along four dimensions: completeness, answer relevancy, faithfulness, and unanswerability. We additionally report Visual Source F1 to directly measure visual citation accuracy. Although proprietary frontier models still achieve the best overall scores on the VinQA test split, fine-tuning open Qwen2.5-VL models on the VinQA training split substantially improves their performance and markedly narrows this gap. Modality Encoding is initially more robust than Page Encoding for complex documents with long text, many visual elements, and diverse visual citation requirements. After training on VinQA, however, Page Encoding reaches a comparable performance level, showing that it can compete effectively even without the explicit parsing used in Modality Encoding. Finally, Visual G-Eval, an MLLM-based judge, confirms that fine-tuned models insert visual elements at semantically appropriate positions with faithful supporting text.

\end{abstract}    
\section{Introduction}
\label{sec:intro}

\begin{figure}
    \centering
    \includegraphics[width=\columnwidth]{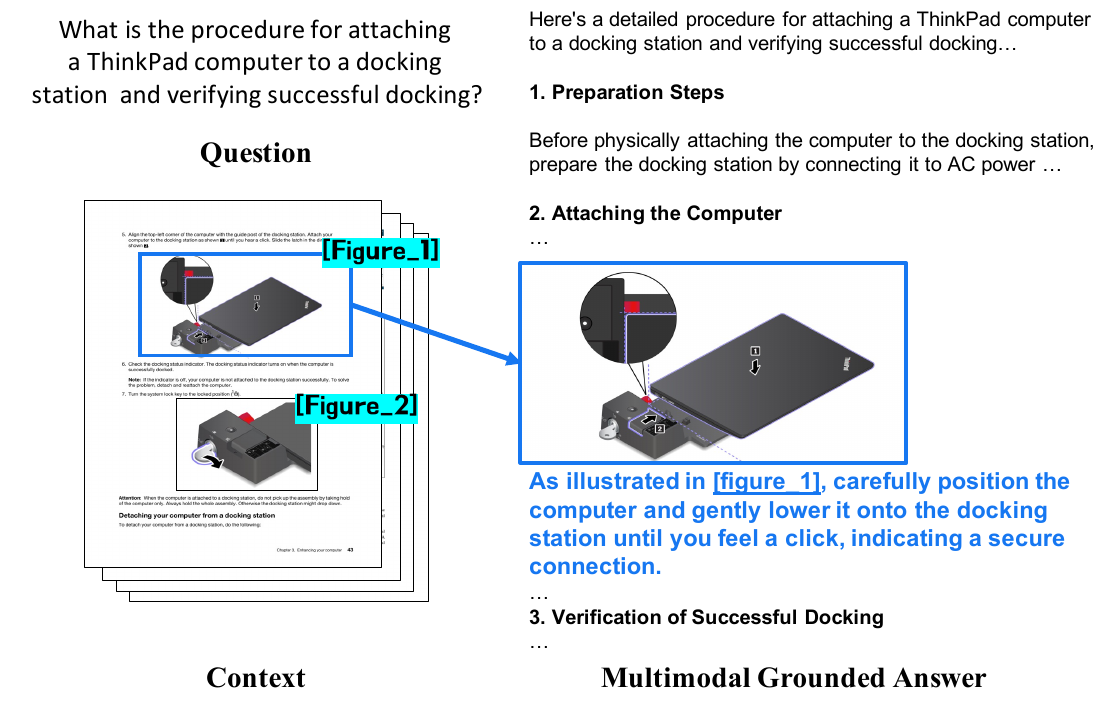}
    \caption{The example of VinQA. Each question is paired with a context, and the multimodal grounded answer cites relevant visual elements with supporting text, inserting each cited element immediately before its citing sentence.}
    \label{fig:overview}
\end{figure}

 
Real-world documents are inherently multimodal, comprising not only text but also various visual elements (e.g., tables, charts, photographs, and diagrams) arranged in diverse layouts. With the advent of Multimodal Large Language Models (MLLMs) \cite{multimodal_survey, qwen2.5-vl, internvl3}, various studies have explored question answering over such documents. Some approaches process entire document pages \cite{ma2025mmlongbench, deng2024longdocurl, ding2024mvqa, zhu2024mmdocbench, m-longdoc}, while others first retrieve relevant pages using a multimodal retriever in a multimodal Retrieval-Augmented Generation (RAG) scenario, and then generate grounded answers from the retrieved context \cite{m3docrag, visdom, visrag}. However, these studies only consider text-only answers, underutilizing the rich visual elements embedded in the documents.

For example, as illustrated in Figure~\ref{fig:overview}, when a question pertains to a specific part of a product’s installation process, providing textual explanations alongside a figure that illustrates the corresponding installation step can significantly enhance understanding. Similarly, figures (e.g., diagrams) are effective for conveying complex ideas in research domains, and charts and tables are well-suited for presenting quantitative information in financial domains.

Motivated by these benefits, some recent work has begun to interleave visual elements into the answers~\cite{yu2025mramg,mcitebench,yan2025m}. However, these approaches typically either operate on narrow-domain document collections or give limited consideration to the procedure for detecting and citing visual regions from raw document page images.

In this paper, we introduce VinQA, the Visual Elements Interleaved Long-form Answer Generation in Question Answering dataset, consisting of training and test splits. VinQA is built from diverse real-world documents that combine text with various visual elements. As shown in Figure~\ref{fig:overview}, each VinQA instance includes (i) a Question requiring both textual and visual understanding, (ii) a Context of the top-$K$ relevant pages obtained through a multimodal RAG pipeline, and (iii) a Multimodal Grounded Answer—a long-form response grounded in the context, where visual elements are interleaved before the supporting text that cites them.



We propose two encoding methods for feeding raw document page images into the MLLM, each with a visual-element citation mechanism. In \textit{Page Encoding}, we directly encode each page as visual tokens together with bounding boxes detected for visual elements, and the model cites these regions via their identifiers. In \textit{Modality Encoding}, we parse each page with OCR to extract text and crop detected visual elements, encode them separately, and cite each cropped element via its identifier. \textit{Page Encoding} preserves full-page layout by operating directly on raw images, but it also requires the model to read text and interpret bounding-box regions from the image itself.



In our experiments, we propose M-GroSE (Multimodal Grounded QA Scoring Evaluator), which extends GroUSE~\cite{grouse} from text-only long-form grounded answers to multimodal answers, and evaluates three dimensions: relevance, completeness, and faithfulness. We also report Visual Source F1, which directly measures visual-element citation accuracy. Although proprietary frontier models still achieve the highest scores on these metrics on the VinQA test split, fine-tuning the open-source Qwen2.5-VL~\cite{qwen2.5-vl} model on the VinQA training split substantially improves its performance and markedly narrows this gap. Modality Encoding is initially more robust than Page Encoding on complex documents, for example those with long text and many visual elements, but after training on VinQA the performance gap largely disappears, showing that Page Encoding can perform competitively even without the explicit parsing used in Modality Encoding. In addition, ablations with Visual G-Eval, an MLLM-based evaluator focused on the visual aspects of cited elements, show that VinQA helps models use cited visual elements more effectively and place them in contextually appropriate positions with faithful supporting text.

Our main contributions are summarized as follows:
\begin{enumerate}
\item \textbf{VinQA dataset.} We present a document-grounded QA dataset in which long-form answers interleave cited visual elements with their supporting text.

\item \textbf{Two encoding strategies with visual-element citation.} \emph{Page Encoding} encodes full-page images with bounding boxes of visual elements and treats these boxed regions as citable units. \emph{Modality Encoding} parses pages and then encodes text and cropped visual elements separately, using these crops as citable units.

\item \textbf{Evaluation framework for multimodal grounded answers.} We propose M-GroSE to evaluate the groundedness of multimodal long-form answers. To independently assess visual-element citations, we report Visual Source F1. We also use Visual G-Eval, an MLLM-based evaluator that directly considers the visual content of cited elements.

\item \textbf{Key findings.} We empirically demonstrate that VinQA effectively improves both the groundedness of multimodal answer generation and the accuracy of visual-element citations. In particular, Page Encoding, when trained on VinQA, reaches performance comparable to Modality Encoding even without explicit parsing. Furthermore, Visual G-Eval shows that VinQA helps models insert visually informative elements at contextually appropriate positions with faithful supporting text.

\end{enumerate}

\section{Related Work}

\subsection{Multimodal Document QA}





Research on multimodal document QA has progressed along two main lines: retrieval-augmented pipelines and full-document processing with MLLMs. The former retrieves relevant document pages using a retriever and then generates answers from the retrieved context~\cite{m3docrag, visdom, huybrechts2025document}. VisRAG~\cite{visrag} and VDocRAG~\cite{tanaka2025vdocrag} further explore grounded short-form answer generation by adopting a \textit{Page Encoding} strategy, in which each page is encoded as an image input to the MLLM. In parallel, another line of work performs QA by feeding entire documents directly into an MLLM without a retrieval step~\cite{ma2025mmlongbench, deng2024longdocurl, ding2024mvqa, zhu2024mmdocbench}. M-LongDoc~\cite{m-longdoc} constructs long-form grounded QA data over documents such as academic papers and financial reports, and adopts a \textit{Modality Encoding} approach. 

While these approaches enable multimodal document understanding via Page Encoding and Modality Encoding, the generated answers remain text-only. Therefore, we build on these two encoding methods and equip each with a visual-element citation mechanism, enabling models to generate long-form answers that explicitly interleave cited visual elements with supporting text. 

\subsection{Multimodal Citation Text Generation}

%

Citation text generation aims to produce answers that explicitly cite the given context. Early studies focus on text-only contexts~\cite{gao-etal-2023-enabling, liu-etal-2023-evaluating, longcite}, while recent work extends the task to multimodal settings~\cite{yan2025m}. VISA~\cite{ma-etal-2025-visa} additionally predicts bounding boxes as visual evidence for the cited source regions. However, VISA does not interleave cited visual regions into the answer with supporting text, as VinQA does.

MCiteBench~\cite{mcitebench} requires citing both textual spans and visual elements (e.g., figures, tables), but its context is limited to pre-extracted text snippets and cropped images of visual-elements from academic papers. MRAMG-Bench~\cite{mramgbench} generates multimodal answers grounded in pre-separated text blocks and images extracted from academic, web, and lifestyle documents. In contrast, we start from raw document page images and investigate how to convert them into MLLM inputs with an explicit citation mechanism. Our dataset also covers a broader range of document domains (Table~\ref{tab:dataset_stats}), and we further introduce M-GroSE, an evaluation framework tailored to assessing such multimodal grounded answers.

\section{Visual Elements Interleaved Long-Form Answer Generation in Question Answering (VinQA) Dataset}
\label{sec:data_gen}

\begin{table*}[t]
\centering
\small
\resizebox{0.9\textwidth}{!}{%
\begin{tabular}{l|ccc|ccc|cc}
\toprule
\multirow{2}{*}{\textbf{Dataset}} &
\multicolumn{3}{c|}{\textbf{Question Type}} &
\multicolumn{3}{c|}{\textbf{Contexts}} &
\multicolumn{2}{c}{\textbf{Answer}} \\
& \textbf{Cross-page} & \textbf{Cross-modal} & \textbf{Unans.} &
\textbf{Multi Doc} & \textbf{Doc Domains} & \textbf{Visual Elements} &
\textbf{Type} & \textbf{Modalities} \\
\midrule
\textbf{SlideVQA~\cite{tanaka2023slidevqa}}                               & \checkmark & \checkmark & \ding{55} & \ding{55} & S             & Ch/Fg/Tb     & Short & \textbf{T} \\

\textbf{MMLongBench\textendash Doc~\cite{ma2025mmlongbench}}             & \checkmark & \checkmark & \checkmark & \ding{55} & P/G/R/F/S   & Ch/Fg/Tb     & Short & \textbf{T} \\
\textbf{LongDocURL~\cite{deng2024longdocurl}}                             & \checkmark & \checkmark & \checkmark & \ding{55}  & P/T/G/R/S             & Ch/Fg/Tb     & Short & \textbf{T} \\
\textbf{VisDoM~\cite{visdom}}                                 & \checkmark & \checkmark & \ding{55} & \checkmark & P/W/S         & Ch/Fg/Tb     & Short  & \textbf{T} \\
\textbf{Document Haystacks \& V\textendash RAG~\cite{huybrechts2025document}} & \ding{55}  & \checkmark & \ding{55} & \checkmark & Etc           & Fg/Tb          & Short & \textbf{T} \\
\textbf{VDocRAG~\cite{tanaka2025vdocrag}}                                & \checkmark & \checkmark & \checkmark  & \checkmark & W/S/Etc & Ch/Fg/Tb     & Short & \textbf{T} \\
\textbf{ViDoRAG~\cite{wang2025vidoragvisualdocumentretrievalaugmented}}                                & \ding{55} & \ding{55} & \ding{55}  & \checkmark & P/W/S & Ch/Fg/Tb     & Short & \textbf{T} \\
\textbf{VisRAG~\cite{visrag}}                                 & \checkmark & \checkmark & \ding{55}  & \checkmark & P/W/T/G/S/Etc & Ch/Fg/Tb     & Short & \textbf{T} \\
\textbf{M\textendash LongDoc~\cite{m-longdoc}}                   & \checkmark & \ding{55} & \ding{55} & \ding{55} & P/G/F         & Ch/Fg/Tb     & Long  & \textbf{T} \\
\textbf{VISA~\cite{ma-etal-2025-visa}}                                 & \ding{55} & \ding{55} & \ding{55} & \checkmark & P/W & Fg/Tb     & Short & \textbf{T+V} \\
\textbf{MCiteBench~\cite{mcitebench}}                             & \ding{55} & \checkmark & \ding{55} & \ding{55} & P             & Fg/Tb     & Long  & \textbf{T+V} \\
\textbf{M\-DocSum~\cite{yan2025m}}                                 & \checkmark & \checkmark & \ding{55} & \ding{55} & P & Fg/Tb     & Long & \textbf{T+V} \\
\textbf{MRAMG\textendash Bench~\cite{yu2025mramg}}                 & \checkmark & \checkmark & \ding{55} & \checkmark & P/W/G/Etc         & Fg/Tb     & Long  & \textbf{T+V} \\
\midrule
\textbf{VinQA (Ours)}                      & \checkmark & \checkmark & \checkmark & \checkmark & P/W/T/G/R/F/S & Ch/Fg/Tb & Long  & \textbf{T+V} \\
\bottomrule
\end{tabular}%
}
\caption{Comparison of multimodal document QA datasets in terms of question type, context, and answer. \textbf{Question Types} include \textbf{Cross-page}, \textbf{Cross-modal}, and \textbf{Unanswerable} questions. \textbf{Contexts} comprise \textbf{Multi Doc}, \textbf{Doc Domain}, and \textbf{Visual Elements}. \textbf{Doc Domains}: \textbf{P} (Academic Paper), \textbf{W} (Web), \textbf{T} (Textbook), \textbf{G} (Guidebook), \textbf{R} (Research Report), \textbf{F} (Financial Report), \textbf{S} (Slides), \textbf{Etc} (Others). \textbf{Visual Elements}: \textbf{Ch} (Chart), \textbf{Fg} (Figure), \textbf{Tb} (Table). \textbf{Answer} consists of \textbf{Type} (Short vs.\ Long) and \textbf{Modalities} (\textbf{T} vs.\ \textbf{T+V}), where \textbf{T} indicates text-only answers, and \textbf{T+V} denotes answers grounded on both textual and visual elements.}
\label{tab:compare_vinqa}
\end{table*}

\label{sec:data_gen}
\begin{figure*}[ht!]
    \centering
    \includegraphics[width=0.9\linewidth]{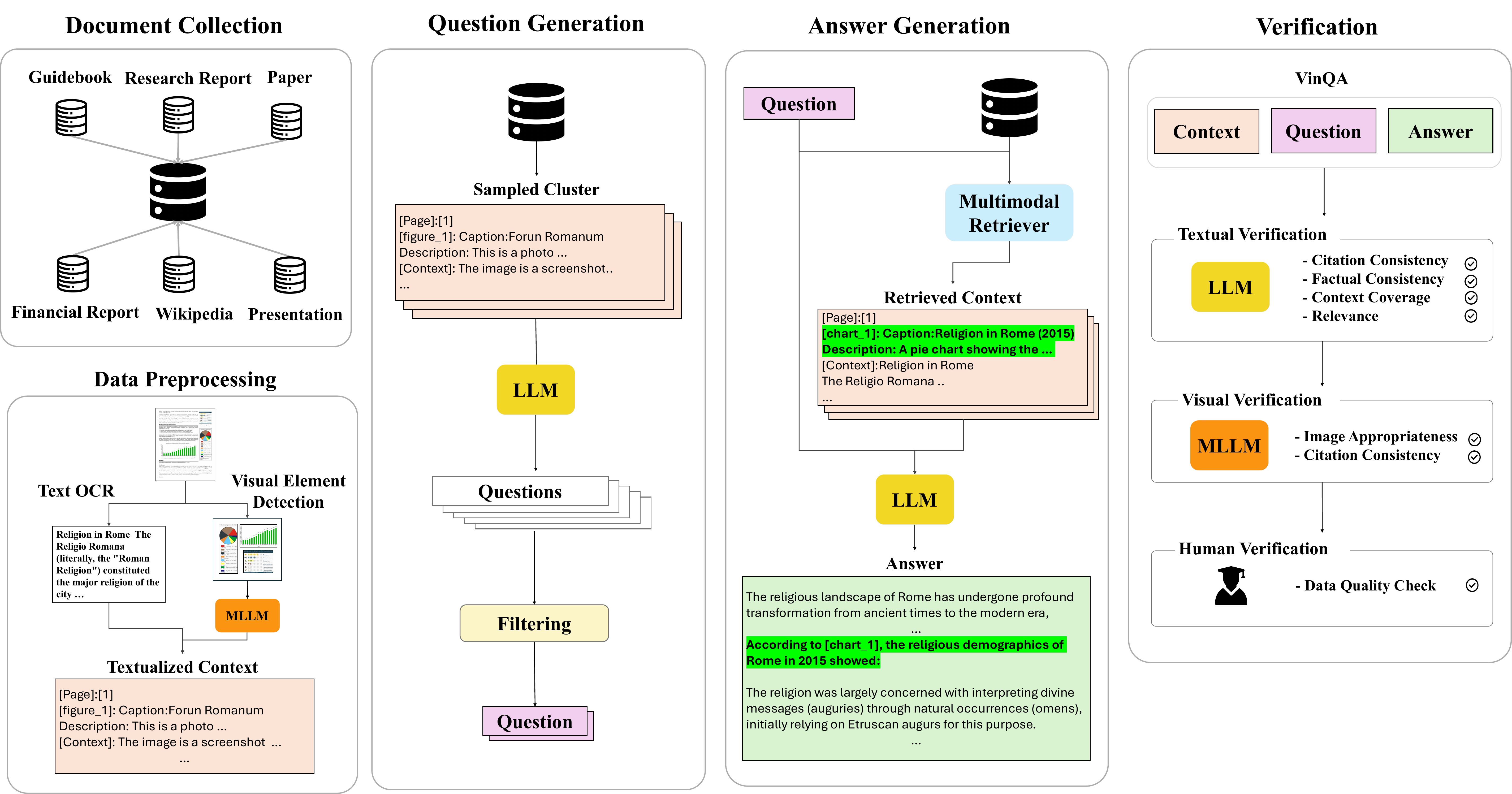}
    \caption{Overview of VinQA construction process. We design the data construction process with the consideration of simulating a Multimodal RAG pipeline over real-world documents.}
    \label{fig:data_gen_pipeline}
\end{figure*}

\begin{table}[ht!]
  \centering
  \small
  \resizebox{0.7\columnwidth}{!}{
  \begin{tabular}{lrr}
    \toprule
    \textbf{Category} & \textbf{Train} & \textbf{Test} \\
    \midrule
    \rowcolor[gray]{0.9}\multicolumn{3}{c}{\textbf{Documents}} \\
    \midrule
    \textbf{Total (\# Pages)} & \textbf{131,906} & \textbf{9,373} \\
    \hspace{1em}Guidebook        & 19,616 & 1,278 \\
    \hspace{1em}Financial report & 1,554  & 1,681 \\
    \hspace{1em}Academic Paper            & 35,331 & 2,081 \\
    \hspace{1em}Presentation     & 48,703 & 1,953 \\
    \hspace{1em}Research report  & 398    & 565 \\
    \hspace{1em}Web        & 16,508 & 1,815 \\
    \hspace{1em}Textbook         & 9,796  & -- \\
    \midrule
    \rowcolor[gray]{0.9}\multicolumn{3}{c}{\textbf{QA}} \\
    \midrule
    \textbf{Total} & \textbf{42,700} & \textbf{1,605} \\
    \hspace{1em}Answerable   & 39,700 & 1,206 \\
    \hspace{1em}Unanswerable & 3,000  & 399 \\
    \midrule
    \multicolumn{3}{l}{\textbf{Answerable Q Types}} \\
    \hspace{1em}Single-page  & 4,030  & 176 \\
    \hspace{1em}Multi-page   & 35,670 & 1,030 \\
    \midrule
    \hspace{1em}Text-only                 & 12,949 & 369 \\
    \hspace{1em}Text + Visual Elements    & 26,751 & 837 \\
    \hspace{2em}- Chart                   & 4,170  & 204 \\
    \hspace{2em}- Figure                  & 9,984  & 226 \\
    \hspace{2em}- Table                   & 7,010  & 293 \\
    \hspace{2em}- Mixed                   & 5,587  & 114 \\
    \bottomrule
  \end{tabular}
  }
  \caption{Statistics of VinQA. Answerable question types are categorized as follows: \textit{Single-page} and \textit{Multi-page} indicate how many pages each question must be grounded in, while \textit{Text-only} and \textit{Text + Visual Elements} indicate which modalities are used for grounding that question.}
  \label{tab:dataset_stats}
\end{table}



We introduce the VinQA dataset for question answering over multimodal contexts, with its statistics summarized in Table~\ref{tab:dataset_stats} and key characteristics compared with prior datasets in Table~\ref{tab:compare_vinqa}.

(i) \textbf{Question.} VinQA covers diverse question types, including cross-page (multi-page reasoning), cross-modal (requiring both textual and diverse types of visual evidence), and unanswerable questions. (ii) \textbf{Context.} The context is built using multimodal retriever over diverse document corpus spanning heterogeneous document domains—academic paper, web pages, textbooks, guidebooks, research reports, financial reports, and slides (P/W/T/G/R/F/S). It contains rich visual elements (charts, tables, figures) and complex layouts (e.g., slides). Moreover, it supports multi-document cases, where a single instance may comprise pages taken from different documents. (iii) \textbf{Answer.} Each answer is a structured, long-form response that includes a header and a body, grounded in both text and visual elements. Cited visual elements are interleaved at contextually appropriate positions with supporting text.


We first use an MLLM to textualize each multimodal document, followed by an LLM that generates corresponding questions, contexts, and answers. The resulting data are organized into multimodal inputs using the encoding methods in Section~\ref{sec:method}, enabling MLLMs to inherit and extend LLM grounding capabilities. The overall construction process is illustrated in Figure~\ref{fig:data_gen_pipeline}.

\subsection{Document Collection and Data Preprocessing}
\label{sec:doc_processing}
To construct the VinQA corpus, we first collect multimodal documents from real-world sources by aggregating the source documents from six public document QA datasets: MMLongBench-Doc \cite{ma2025mmlongbench}, TAT-DQA \cite{zhu2022towards}, VisDoM \cite{visdom}, SlideVQA \cite{tanaka2023slidevqa}, MMDocIR \cite{dong2025mmdocir}, and VisRAG \cite{visrag}. We only utilize the raw documents from these datasets and do not use their original question–answer annotations. We manually classify the documents into seven domains based on topic and format, as shown in Table~\ref{tab:compare_vinqa}. The corpus is split into training and test sets using domain-balanced stratified sampling, ensuring document-level separation.\footnote{We exclude textbook documents from the test split because its pages, unlike other categories, lack clear document boundaries, making it impossible to avoid overlap between training and test corpus.}

To textualize document pages, we extract text using OCR and convert visual elements into structured textual representations. Concretely, we detect charts, tables, and figures with DocLayout~\cite{doclayout} to obtain their bounding boxes on each page. For every detected element, we prompt an MLLM with two  types of inputs: (i) the full page image, where the target element is highlighted with a bounding box, and (ii) the cropped image of the target element. Using these inputs, the MLLM generates two textual outputs: (a) a caption-the in-document text that directly refers to the visual element, and (b) a description-a longer explanation derived from surrounding text and the visual features of the element itself.

\subsection{Question and Answer Generation}

We generate questions that are naturally asked about our collected document corpus. Specifically, we randomly sample a document page, extract its image embedding using ColPali~\cite{colpali}, and cluster pages by embedding similarity. An LLM then generates questions for each cluster, including cases that require reasoning across multiple pages and modalities, thereby encouraging multi-page and cross-modal grounding in subsequent QA tasks. Following VisRAG~\cite{visrag}, we discard questions that rely on local context but omit explicit entity names, as such questions are ambiguous for retrieval.


We then use ColPali to retrieve the top-$K$ most relevant document pages for each question; these pages serve as the context. During answer generation, we prompt the LLM with this context to produce a long-form grounded answer that cites visual elements via their identifiers, along with faithful supporting text at contextually appropriate positions. In a post-processing step, each cited visual element is inserted immediately above the paragraph where its identifier appears.

To construct unanswerable QA examples, we first form hard negatives by retrieving lower-ranked pages from ColPali that have little or no relevance to the given question. We then retain only those cases where the LLM correctly identifies the question as unanswerable. This procedure mimics realistic retrieval errors that may occur in practice and makes the task more challenging by requiring the model to distinguish between answerable and unanswerable questions based on the retrieved content.

\subsection{Data Verification}

To ensure the quality of the generated QA data, we perform a three-step verification process consisting of textual, visual, and human verification. For the textual verification step, each question–context–answer triple is evaluated by a state-of-the-art LLM along four key aspects: citation accuracy, factual consistency with the context, completeness of relevant information, and absence of irrelevant content.
For the visual verification step, a state-of-the-art MLLM checks whether visual elements (e.g., charts, tables, figures) are properly cited and semantically aligned with the corresponding statements.

For the test set, we further reduce noise by running an additional textual verification pass with a different LLM and by performing human verification focused on visual aspects. A team of four human annotators manually inspects each QA pair following a structured guideline that checks (i) whether the bounding boxes of visual elements are correctly specified, (ii) whether visual citations are accurate and necessary to support the answer, and (iii) whether the supporting text is free of hallucinations. After human verification, 106 answerable and one unanswerable QA pairs were filtered out, resulting in a highly reliable and clean test set.

\section{Methods}
\label{sec:method}
\begin{figure*}[ht!]
    \centering
    \includegraphics[width=0.9\linewidth]{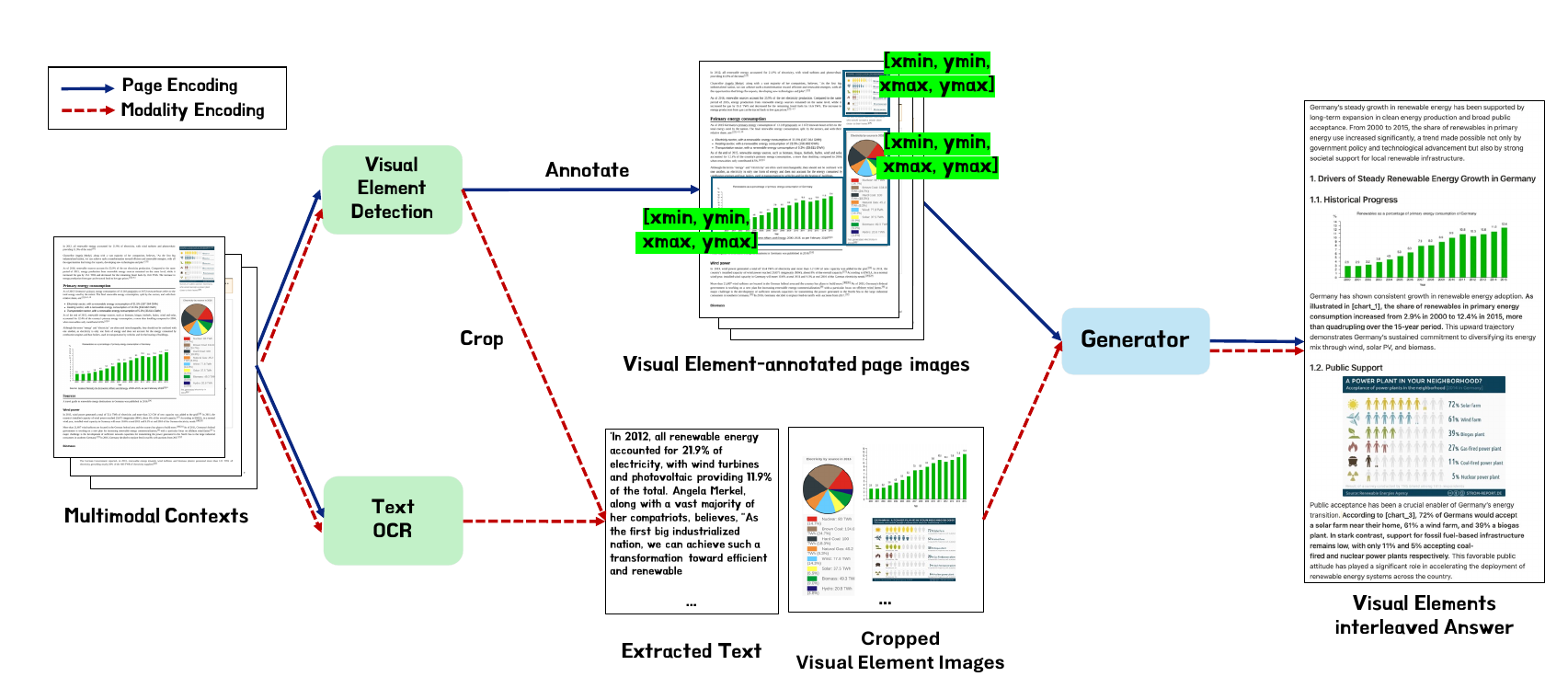}
    \caption{Visual Elements Interleaved Answer Generation based on encoding methods. The \textcolor{blue}{blue} solid arrows represent the \textit{Page Encoding} method, and the \textcolor{red}{red} dotted arrows represent the \textit{Modality Encoding} method.}
    \label{fig:encoding_methods}
\end{figure*}

Given a user query $q$ and a context composed of $n$ document page images $\mathcal{P} = \{p_1, p_2, \dots, p_n\}$, the MLLM is tasked with producing a grounded long-form answer $A = \{x_1, x_2, \dots, x_k\}$, where each $x_i$ is a text span. Some spans include a specific visual-element identifier token for citation, and the corresponding visual element is interleaved immediately preceding the span. This section describes two encoding methods for converting $\mathcal{P}$ into MLLM inputs and enabling citation of visual elements: \textit{Page Encoding} and \textit{Modality Encoding}. Figure~\ref{fig:encoding_methods} illustrates the overall workflow with these two encoding methods.


\subsection{Page Encoding}


Inspired by VisRAG~\cite{visrag}, we directly use document page images $\mathcal{P}$ as model input. Additionally, we use DocLayout~\cite{doclayout} to detect visual elements on each page. The detected elements are annotated with bounding boxes, which are provided to the MLLM alongside the page image as auxiliary input.

The processed input is represented as:
\[
\left\{ \left( p_i, \text{BBoxList}_i \right) \right\}_{i=1}^n
\]
where $p_i$ denotes the page image of the $i$-th document page, which is encoded into visual tokens, and $\text{BBoxList}_i = \{ b_i^{(1)}, b_i^{(2)}, \dots \}$ is the set of bounding box coordinates for visual elements in $p_i$, each encoded into text tokens.  Each bounding box is assigned a unique visual element identifier, allowing the MLLM to cite the corresponding visual element during answer generation.

This input preserves all visual information present in the page image (e.g., layout). However, it imposes additional challenges for the MLLM: it must (i) interpret textual content directly from pixels (without explicit OCR) and (ii) align each visual-element identifier with its bounding box region on the page for accurate understanding and citation.


\subsection{Modality Encoding}

Unlike \textit{Page Encoding}, we first extract text from document images via OCR and detect visual element regions and crop them according to their bounding boxes. The extracted text is then encoded into text tokens, and the cropped visual element images are encoded into visual tokens.

The   input is represented as:
\[
\left\{ \left(  t_i, \text{V}_i \right) \right\}_{i=1}^n
\]
where $t_i$ denotes the extracted text from the $i$-th page, and $\text{V}_i = \{ v_i^{(1)}, v_i^{(2)}, \dots \}$ is the set of visual element crops derived from $p_i$. Each crop is assigned a unique visual-element identifier, which is used to cite the corresponding visual element during answer generation.

While this approach loses some visual information (e.g., layout) during processing, it enables fine-grained understanding by processing text and visual elements independently with modality-specific representations.



\section{Experiments}


\subsection{Main Results}



\begin{table*}[ht!]
  \centering
  \footnotesize
  \begin{tabular}{l c c c c c c}
    \toprule
    \multirow{2}{*}{\textbf{Model}}
      & \multicolumn{5}{c}{\textbf{M-GroSE}}
      & \textbf{Visual Source} \\
    \cmidrule(lr){2-6}
      & \textbf{Relevance}
      & \textbf{Completeness}
      & \textbf{Faithfulness}
      & \textbf{Unans. (F1)}
      & \textbf{Avg}
      & \textbf{F1} \\
    \midrule

    \rowcolor{gray!25}
    \multicolumn{7}{c}{\itshape\textbf{Page Encoding}} \\
    \midrule
    \textbf{Proprietary Models} 
    \\
    GPT-4.1
      & \textbf{4.88} &  \underline{3.68} & \underline{4.27} & 0.78 & \underline{3.46}
      & \underline{0.62} \\
    GPT-4.1-mini
      & 4.57 & 3.58 & 4.00 & 0.47 & 3.29
      & 0.49 \\
    Gemini 2.0 Flash
      & 4.19 & 2.60 & 3.62 & \underline{0.86} & 2.86
      & 0.46 \\
    Claude 3.5 Sonnet
      & \underline{4.83} & \textbf{3.81} & \textbf{4.46} & 0.70 & \textbf{3.53}
      & \textbf{0.65} \\
    \midrule
    \textbf{Open-source Models} \\
    InternVL3-8B
      & 3.26 & 1.66 & 2.24 & 0.78 & 2.05
      & 0.23 \\
    Qwen2.5-VL-7B
      & 3.01 & 1.61 & 2.32 & 0.71 & 1.99
      & 0.31 \\
    \rowcolor{cyan!15}
    Qwen2.5-VL-7B (VinQA)
      & 4.60 & 3.65 & 4.09 & \textbf{0.91} & 3.34
      & 0.55 \\
   \midrule
    \rowcolor{gray!25}
    \multicolumn{7}{c}{\itshape\textbf{Modality Encoding}} \\
    \midrule
    \textbf{Proprietary Models} 
    \\
    GPT-4.1
      & \textbf{4.89} & \textbf{3.98} & \underline{4.60} & 0.70  & \textbf{3.63} {\tiny\textcolor{blue}{(+0.16)}}
      & \textbf{0.72} {\tiny\textcolor{blue}{(+0.10)}} \\
    GPT-4.1-mini
      & 4.63 & 3.94 & 4.22 & 0.76 & 3.45 {\tiny\textcolor{blue}{(+0.16)}}
      & 0.60 {\tiny\textcolor{blue}{(+0.11)}} \\
    Gemini 2.0 Flash
      & 4.65 & 3.51 & 4.52 & \underline{0.86} & 3.43 {\tiny\textcolor{blue}{(+0.57)}}
      & 0.61 {\tiny\textcolor{blue}{(+0.14)}} \\
    Claude 3.5 Sonnet
      & \underline{4.83} & \underline{3.94} & \textbf{4.68} & 0.69  & \underline{3.62} {\tiny\textcolor{blue}{(+0.09)}}
      & \underline{0.70} {\tiny\textcolor{blue}{(+0.06)}} \\
    \midrule
    \textbf{Open-source Models} \\
    InternVL3-8B
      & 3.48 & 1.89 & 2.57 & 0.77 & 2.24 {\tiny\textcolor{blue}{(+0.19)}}
      & 0.32 {\tiny\textcolor{blue}{(+0.09)}} \\
    Qwen2.5-VL-7B
      & 2.96 & 1.81 & 2.53 & 0.75 & 2.08 {\tiny\textcolor{blue}{(+0.09)}}
      & 0.40 {\tiny\textcolor{blue}{(+0.09)}} \\
    \rowcolor{cyan!15}
    Qwen2.5-VL-7B (VinQA)
      & 4.59 & 3.59 & 4.09 & \textbf{0.90} & 3.33 {\tiny\textcolor{red}{(-0.02)}}
      & 0.58 {\tiny\textcolor{blue}{(+0.02)}} \\
    \bottomrule
  \end{tabular}
  \caption{Overall performance on the VinQA test set. The values in parentheses show the difference between \textit{Modality Encoding} and \textit{Page Encoding}. The M-GroSE Avg score is computed as the mean of the three 1–5 criteria (relevance, completeness, and faithfulness) and the unanswerability F1 score, where the latter is linearly rescaled from [0,1] to [1,5] before averaging.}
   
  \label{table:main_result}
\end{table*}

 \textbf{Implementations} 
We adopt Qwen2.5-VL-7B~\cite{qwen2.5-vl} and train for 3 epochs with both encoding methods on the VinQA training split, conducted on 16×A100 GPUs.

\noindent \textbf{Evaluation} We propose Multimodal Grounded QA Scoring Evaluator (M-GroSE), a modified version of GroUSE
~\cite{grouse} originally designed for text-only long-form grounded answers, extended to assess multimodal grounding by incorporating visual citations.
For answerable questions, we use GPT-5-mini as an LLM-judge to evaluate answers along three criteria: relevance (whether the answer addresses the question), completeness (whether it includes all necessary information from the retrieved context), and faithfulness (whether it is grounded in the sources without hallucination). For unanswerable questions, we report Unanswerability F1, which measures whether the model correctly abstains when no valid answer exists in the provided context. To extend GroUSE to multimodal QA, we textualize both textual and visual context as done in Section~\ref{sec:doc_processing} and modify the LLM-judge prompt to evaluate both textual reasoning and visual citations with their supporting text. 

In addition, we evaluate visual citation accuracy using Visual Source F1, following MCiteBench~\cite{mcitebench}, by comparing predicted visual-element references with VinQA’s gold citations.
\noindent \textbf{Overall Performance} As shown in Table~\ref{table:main_result}\footnote{We evaluate recent state-of-the-art MLLMs, but exclude models that rely on test-time scaling for a fair comparison}, on the VinQA test split, the best Page Encoding performance is achieved by Claude 3.5 Sonnet, with M-GroSE Avg of 3.53 and Visual Source F1 of 0.65, while the best Modality Encoding result comes from GPT-4.1 (M-GroSE Avg of 3.63, Visual Source F1 of 0.72), showing a clear gap between proprietary and open-source models.
However, fine-tuning Qwen2.5-VL-7B on VinQA significantly improves both metrics; fine-tuned model reaches 3.34 (M-GroSE Avg), 0.55 (Visual Source F1) with Page Encoding and 3.33 (M-GroSE Avg), 0.58 (Visual Source F1) with Modality Encoding, markedly narrowing the gap to the SOTA. These results suggest that VinQA provides highly effective supervision for multimodal grounding and visual-element citation.


Notably, \textit{Modality Encoding} generally outperforms \textit{Page Encoding} on both M-GroSE Avg and Visual Source F1—as indicated by the positive deltas in parentheses in Table~\ref{table:main_result}—except for models fine-tuned on VinQA. This indicates that while Modality Encoding is more effective prior to fine-tuning, Page Encoding can achieve comparable performance once fine-tuned on the dataset. We analyze this result in detail in Section~\ref{sec:analysis}.

\subsection{Analysis}\label{sec:analysis}

\begin{figure*}[ht!]
\centering
\begin{subfigure}[b]{0.26\textwidth}
\centering
\begin{tikzpicture}
\begin{axis}[
    title={Modality Enc},
    width=1.0\linewidth,
    height=3.2cm,
    xlabel={Context Token Length\\[4pt]
            \makebox[0.9\linewidth][c]{\footnotesize (a)}},
    xlabel style={align=center,font=\scriptsize},
    ylabel={M-GroSE Avg},
    ylabel style={font=\scriptsize, yshift=-1em},
    title style={font=\scriptsize},
    xmin=0, xmax=12500, ymin=0, ymax=4,
    xtick={0,2500,5000,7500,10000,12500},
    xticklabels={0,2.5k,5k,7.5k,10k,MAX},
    scaled x ticks=false,
    tick scale label code/.code={},
    tick label style={font=\scriptsize},
    extra y ticks={0},
    extra y tick style={black,line width=1pt},
    ymajorgrids=true,
    grid style=dashed,
    legend style={
        legend columns=1,
        font=\scriptsize,
        /tikz/every even column/.style={column sep=0.8em}
    },
    legend cell align={left},
    legend to name=commonlegend2,
]
\addplot[color=blue!60,mark=triangle*,thick]
  coordinates {(1250,2.17) (3750,1.93) (6250,2.10) (8750,2.12) (11000,2.16)};
\addlegendentry{\scriptsize Qwen2.5-VL 7B}

\addplot[color=red!60,mark=square*,thick]
  coordinates {(1250,3.42) (3750,3.46) (6250,3.41) (8750,3.15) (11000,3.03)};
\addlegendentry{\scriptsize Qwen2.5-VL 7B (VinQA)}
\end{axis}
\end{tikzpicture}
\end{subfigure}%
\hfill%
\begin{subfigure}[b]{0.26\textwidth}
\centering
\begin{tikzpicture}
\begin{axis}[
    title={Page Enc},
    width=1.0\linewidth,
    height=3.2cm,
    xlabel={Context Token Length\\[4pt]
            \makebox[0.9\linewidth][c]{\footnotesize (b)}},
    xlabel style={align=center,font=\scriptsize},
    ylabel={M-GroSE Avg},
    ylabel style={font=\scriptsize, yshift=-1em},
    title style={font=\scriptsize},
    xmin=0, xmax=12500, ymin=0, ymax=4,
    xtick={0,2500,5000,7500,10000,12500},
    xticklabels={0,2.5k,5k,7.5k,10k,MAX},
    scaled x ticks=false,
    tick scale label code/.code={},
    tick label style={font=\scriptsize},
    ymajorgrids=true,
    grid style=dashed,
    extra y ticks={0},
    extra y tick style={black,line width=1pt},
]
\addplot[color=blue!60,mark=triangle*,thick]
  coordinates {(1250,2.27) (3750,1.95) (6250,2.04) (8750,1.97) (11000,1.69)};
\addplot[color=red!60,mark=square*,thick]
  coordinates {(1250,3.40) (3750,3.50) (6250,3.43) (8750,3.22) (11000,2.98)};
\end{axis}
\end{tikzpicture}
\end{subfigure}%
\hfill%
\begin{subfigure}[b]{0.26\textwidth}
\centering
\begin{tikzpicture}
\begin{axis}[
    title={Modality Enc - Page Enc (Difference)},
    width=1.0\linewidth,
    height=3.2cm,
    xlabel={Context Token Length\\[4pt]
            \makebox[0.9\linewidth][c]{\footnotesize (c)}},
    xlabel style={align=center,font=\scriptsize},
    ylabel={M-GroSE avg},
    ylabel style={font=\scriptsize, yshift=-1em},
    title style={font=\scriptsize},
    xmin=0, xmax=12500, ymin=-0.2, ymax=0.6,
    xtick={0,2500,5000,7500,10000,12500},
    xticklabels={0,2.5k,5k,7.5k,10k,MAX},
    scaled x ticks=false,
    tick scale label code/.code={},
    tick label style={font=\scriptsize},
    ymajorgrids=true,
    grid style=dashed,
    extra y ticks={0},
    extra y tick style={black,line width=1pt},
]
\addplot[color=blue!60,mark=triangle*,thick]
  coordinates {(1250,-0.10) (3750,-0.03) (6250,0.06) (8750,0.15) (11000,0.47)};
\addplot[color=red!60,mark=square*,thick]
  coordinates {(1250,0.02) (3750,-0.04) (6250,-0.02) (8750,-0.07) (11000,0.05)};
\end{axis}
\end{tikzpicture}
\end{subfigure}%
\hfill%
\raisebox{1.3cm}{%
\pgfplotslegendfromname{commonlegend2}%
}

\caption{M-GroSE Avg performance across context token length.}
\label{fig:grose_context_length}
\end{figure*}
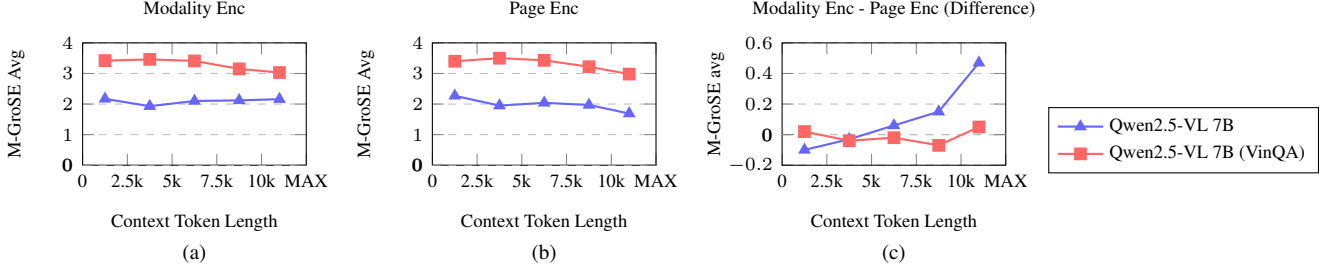

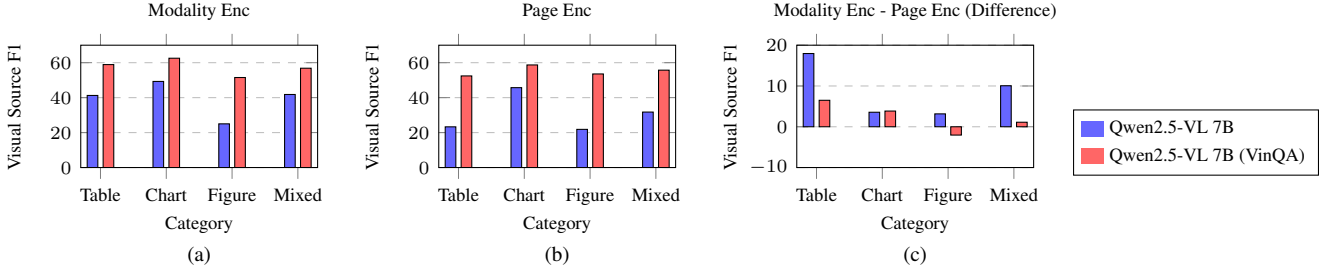
\begin{figure*}[htbp]
\centering
\begin{subfigure}[b]{0.27\textwidth}
\centering
\begin{tikzpicture}
\begin{axis}[
    title={Modality Enc},
    ybar,
    bar width=4pt,
    width=1.0\linewidth,
    height=3.2cm,
    xlabel={Category\\[4pt]\makebox[0.9\linewidth][c]{\footnotesize (a)}},
    ylabel={Visual Source F1},
    symbolic x coords={Table, Chart, Figure, Mixed},
    xtick=data,
    tick label style={font=\scriptsize},
    xlabel style={font=\scriptsize,align=center},
    ylabel style={font=\scriptsize, yshift=-1em},
    title style={font=\scriptsize},
    ymin=0,
    ymax=70,
    legend style={
        legend columns=1,
        font=\scriptsize,
        /tikz/every even column/.style={column sep=0.8em}
    },
    legend cell align={left},
    legend image code/.code={
        \draw[#1, draw=none] (0cm,-0.1cm) rectangle (0.3cm,0.1cm);
    },
    legend to name=commonlegend,
    ymajorgrids=true,
    grid style=dashed,
]
\addplot[fill=blue!60] coordinates {
    (Table, 41.25) (Chart, 49.27) (Figure, 25.00) (Mixed, 41.80)
};
\addplot[fill=red!60] coordinates {
    (Table, 58.91) (Chart, 62.55) (Figure, 51.51) (Mixed, 56.83)
};
\legend{\scriptsize Qwen2.5-VL 7B,\scriptsize Qwen2.5-VL 7B (VinQA)}
\end{axis}
\end{tikzpicture}
\end{subfigure}%
\hfill%
\begin{subfigure}[b]{0.27\textwidth}
\centering
\begin{tikzpicture}
\begin{axis}[
    title={Page Enc},
    ybar,
    bar width=4pt,
    width=1.0\linewidth,
    height=3.2cm,
    xlabel={Category\\[4pt]\makebox[0.9\linewidth][c]{\footnotesize (b)}},
    ylabel={Visual Source F1},
    symbolic x coords={Table, Chart, Figure, Mixed},
    xtick=data,
    tick label style={font=\scriptsize},
    xlabel style={font=\scriptsize,align=center},
    ylabel style={font=\scriptsize, yshift=-1em},
    title style={font=\scriptsize},
    ymin=0,
    ymax=70,
    ymajorgrids=true,
    grid style=dashed,
]
\addplot[fill=blue!60] coordinates {
    (Table, 23.29) (Chart, 45.71) (Figure, 21.84) (Mixed, 31.74)
};
\addplot[fill=red!60] coordinates {
    (Table, 52.42) (Chart, 58.70) (Figure, 53.55) (Mixed, 55.72)
};
\end{axis}
\end{tikzpicture}
\end{subfigure}%
\hfill%
\begin{subfigure}[b]{0.27\textwidth}
\centering
\begin{tikzpicture}
\begin{axis}[
    title={Modality Enc - Page Enc (Difference)},
    ybar,
    bar width=4pt,
    width=1.0\linewidth,
    height=3.2cm,
    xlabel={Category\\[4pt]\makebox[0.9\linewidth][c]{\footnotesize (c)}},
    ylabel={Visual Source F1},
    symbolic x coords={Table, Chart, Figure, Mixed},
    xtick=data,
    tick label style={font=\scriptsize},
    xlabel style={font=\scriptsize,align=center},
    ylabel style={font=\scriptsize, yshift=-1em},
    title style={font=\scriptsize},
    ymin=-10,
    ymax=20,
    ymajorgrids=true,
    grid style=dashed,
]
\addplot[fill=blue!60] coordinates {
    (Table, 17.96) (Chart, 3.56) (Figure, 3.16) (Mixed, 10.06)
};
\addplot[fill=red!60] coordinates {
    (Table, 6.49) (Chart, 3.85) (Figure, -2.04) (Mixed, 1.11)
};
\end{axis}
\end{tikzpicture}
\end{subfigure}%
\hfill%
\raisebox{1.3cm}{%
\pgfplotslegendfromname{commonlegend}%
}

\caption{Visual source F1 across various visual element types}
\label{fig:visual_source_f1_ve_types}
\end{figure*}

\subsubsection*{Does VinQA Improve Model Performance Across Context Token Length and Visual Element Diversity?}


First, we analyze how VinQA fine-tuning affects M-GroSE scores of Qwen2.5-VL-7B with respect to context token length. Figure~\ref{fig:grose_context_length} (a) and (b) show the results for Modality Encoding and Page Encoding, respectively.  We measure context token length under Modality Encoding, which scales with both textual volume and the number or resolution of visual elements on a page. It serves as a proxy for document complexity. Across all context-length bins (0–2.5k to $\ge$10k), VinQA fine-tuning consistently improves multimodal grounded answer quality. However, performance decreases for very long contexts ($\ge$7.5k), suggesting that increased document complexity limits complete and faithful information integration.

Second, we analyze the gains in Visual Source F1 with respect to visual element types in context before and after VinQA fine-tuning. Figure~\ref{fig:visual_source_f1_ve_types} (a) and (b) show the results for Modality Encoding and Page Encoding, respectively. For this analysis, we use the VinQA’s visual-element annotations—Chart, Table, and Figure\footnote{The Figure category includes photos, diagrams, and other non-chart/table visual elements.}—to evaluate performance. 
The ``Mixed'' type covers answers required to include citations to two or more types of visual elements. Before fine-tuning, answers citing Figures perform worst among all types under Modality Encoding, while under Page Encoding both Figures and Tables lag behind the other types. After fine-tuning on VinQA, these categories improve substantially, narrowing their performance gap with the rest. Gains are also observed in the Mixed category, indicating that VinQA enhances citation accuracy even for answers that integrate multiple types of visual modalities.

\subsubsection*{Which is Better: Page Encoding or Modality Encoding?}

We first compare the M-GroSE Avg score difference between \textit{Modality Encoding} and \textit{Page Encoding} across the context token length as shown in Figure~\ref{fig:grose_context_length} (c). Before VinQA fine-tuning, Page Encoding performs slightly better for short contexts ($\le$2.5k), but Modality Encoding yields higher scores as context token length increases. This trend arises because Modality Encoding processes extracted text and cropped visual elements directly, whereas Page Encoding requires the model to interpret textual content from rendered page images and to reason over bounding-box–annotated visual regions. As the context token length increases, this burden on the model grows accordingly. After VinQA fine-tuning, however, this gap becomes negligible across all context lengths. This suggests that Page Encoding enables the model to generate multimodal grounded answers for complex contexts at a level comparable to Modality Encoding, despite not relying on explicit parsing.



We next compare Visual Source F1 difference between \textit{Modality Encoding} and \textit{Page Encoding} across visual element types, as shown in Figure~\ref{fig:visual_source_f1_ve_types} (c). Before VinQA fine-tuning, Modality Encoding achieves higher citation performance for Table and Mixed category. This is likely because tables contain more text than charts or figures, and Mixed requires referencing heterogeneous visual elements; representing cropped regions as separate visual tokens is therefore advantageous. After VinQA fine-tuning, these type-wise gaps narrow considerably, indicating that VinQA helps Page Encoding handle diverse visual elements more effectively, even when operating on page images with bounding-box regions.

\subsubsection*{Are the visual elements appropriately interleaved in the answer with faithful supporting text?}


We have presented evaluation results using M-GroSE and Visual Source F1, but because these metrics operate on textualized context and answers, they are limited in assessing visual aspects. To directly evaluate visual factors, we introduce Visual G-Eval, which uses GPT-5 to score the question and generated answer while consuming the cited visual elements as image inputs. We score each citation on three 1–5 metrics: (1) Effectiveness – how well the cited image, together with its supporting text, visually supports the answer; (2) Position – how appropriately the image is placed within the answer; and (3) Faithfulness – how accurately the supporting text reflects the visual content.

Table~\ref{table:image_citation_g_eval} reports average scores over all visual-element citations across 200 instances sampled from the VinQA test split. Training with VinQA yields consistent gains on all three criteria for both encoding methods, indicating that VinQA provides an effective training signal for learning to better leverage visual elements. Interestingly, Modality Encoding attains relatively high Effectiveness and Faithfulness scores both before and after training. While VinQA training allows Page Encoding to match Modality Encoding in overall multimodal grounded answer performance as measured by M-GroSE, Modality Encoding still retains an advantage in how effectively it uses and explains visual elements to support the answer.

\begin{table}[ht!]
  \centering
  \footnotesize
  \setlength{\tabcolsep}{4pt}  
  \resizebox{0.9\linewidth}{!}{
  \begin{tabular}
  {%
    l                          
   c  c  c                   
  }
    \toprule
    \multirow{2}{*}{\textbf{Model}} 
      & \multicolumn{3}{c}{\textbf{Visual G-Eval}} \\
    \cmidrule(lr){2-4}
      & \textbf{Effectiveness}
      & \textbf{Position}
      & \textbf{Faithfulness} \\
    \midrule

    \rowcolor[gray]{0.9}\multicolumn{4}{c}{\textit{\textbf{Page Encoding}}} \\
    \midrule
    Qwen2.5-VL-7B           & 3.44 & 4.24 & 3.44 \\ 
    \rowcolor{cyan!15}
    Qwen2.5-VL-7B (VinQA)        & \textbf{3.94} & \textbf{4.77} & \textbf{3.85} \\ 

    \midrule
    \rowcolor[gray]{0.9}\multicolumn{4}{c}{\textit{\textbf{Modality Encoding}}} \\
    \midrule
    Qwen2.5-VL-7B           & 4.00 
                        & 4.57 
                        & 4.01 
                        \\ 
                        \rowcolor{cyan!15}
    Qwen2.5-VL-7B (VinQA)          & \textbf{4.17} 
                        & \textbf{4.74} 
                        & \textbf{4.23} 
                        \\ 
    \bottomrule
  \end{tabular}}
  \caption{Visual G-Eval results of Qwen2.5-VL-7B before and after VinQA fine-tuning under Page Encoding and Modality Encoding.}
  \label{table:image_citation_g_eval}
\end{table}

\section{Conclusion}

We introduce the VinQA dataset for generating visual-element–interleaved long-form answers over real-world multimodal documents, together with Page/Modality Encoding and multimodal evaluation methods (M-GroSE, Visual Source F1, Visual G-Eval). Fine-tuning open Qwen2.5-VL models on VinQA training split substantially improves multimodal grounding and makes both encoding strategies competitive, with Page Encoding in particular achieving strong gains even without explicit parsing. As future work, we plan to apply VinQA to models with test-time reasoning (e.g., long think-token generation) and to mitigate performance degradation on more complex, longer-context documents.

\clearpage
{
    \small
    \bibliographystyle{ieeenat_fullname}
    \bibliography{main}
}

\clearpage
\appendix
\setcounter{secnumdepth}{2} 
\renewcommand{\thesubsection}{\thesection.\arabic{subsection}}
\onecolumn

\section{Evaluation Model Choices}
\label{sec:eval_models}

We evaluate state-of-the-art proprietary models—GPT-4.1, GPT-4.1-mini~\cite{gpt4-1}, Gemini 2.0 Flash~\cite{gemini-flash}, and Claude 3.5 Sonnet~\cite{claude3_5}—as well as recent open-source models including InternVL-3~\cite{internvl3} and Qwen2.5-VL~\cite{qwen2.5-vl}. Our evaluation is limited to non-thinking models, excluding models such as o3 and Claude 3.7, which may benefit from additional reasoning steps.

\section{Hyperparameters}
\label{sec:hyperparameters}
We train both the Page Encoding and Modality Encoding variants for three epochs using distributed data-parallel training on 16 × A100 40GB GPUs, which takes 27 and 17 hours, respectively. Table~\ref{tab:my_label} summarizes the training hyperparameters. The two encodings use different maximum image resolutions (measured in total input pixels); in each case, we choose the largest resolution that can be processed on 40GB A100s without out-of-memory errors. Because Modality Encoding only feeds cropped visual elements, it can afford a higher effective resolution for each cropped region, whereas Page Encoding operates on full-page images, so even with a higher overall page image resolution, the effective resolution available for each visual element can be lower.

\begin{table}[ht!]
    \centering
    \begin{tabular}{lrr}
        \toprule
        \textbf{Configuration} & \textit{\textbf{Page Encoding}} & \textit{\textbf{Modality Encoding}} \\
        \midrule
        Epoch & 3 & 3 \\
        Optimizer & AdamW & AdamW \\
        Learning Rate & 1e-05 & 1e-05 \\
        Learning Rate Scheduler & cosine & cosine \\
        Warm-up Ratio & 0.1 & 0.1 \\
        Global Batch Size & 16 & 16 \\
        Grad Acc Steps & 16 & 16 \\
        Numerical Precision & bfloat16 & bfloat16 \\
        Image Resolution & 2508800 & 1003520 \\
        \bottomrule
    \end{tabular}
    \caption{Hyperparameters for training Qwen2.5-VL on VinQA.}
    \label{tab:my_label}
\end{table}

\section{Details of VinQA dataset construction}
\label{sec:appendix_data}



\subsection{Visual-element to textual form transformation}
\label{sec:appendix_data_prepro}
In the data preprocessing process, we generate class, caption, and description for the visual elements in the document page using GPT-4o. Each visual element is classified into categories such as chart, table, photo, diagram, icon, etc. In our dataset construction, we group photo and diagram under the type label ``figure" and exclude elements classified as ``icon" and ``etc". GPT-4o receives two images as input: one containing the full-page image marked with a red bounding box around the target visual element, and the other containing the cropped image of the target visual element. The corresponding prompt is presented in Figure \ref{fig:modal_desc_cap_prompt}.

\subsection{Question Generation}
To generate diverse and domain-balanced queries, we first sample page images uniformly across all domains in the corpus and assign each as a reference page. For every reference page, we utilize a ColQwen\footnote{https://huggingface.co/vidore/colqwen2-v1.0}, multimodal retriever, to gather the ten most visually and semantically similar pages within the corpus. From these ten, four pages are randomly selected and combined with the reference page, yielding a five-page cluster. Each cluster thus encompasses a coherent but non-redundant context for question generation. From the 131,906 page train corpus, 30,000 reference pages are selected, and an equal number of clusters are consequently constructed; in parallel, 2,000 clusters are constructed from the 9,373 page test corpus.

\begin{figure*}[ht!]
    \centering
    \includegraphics[width=0.8\linewidth]{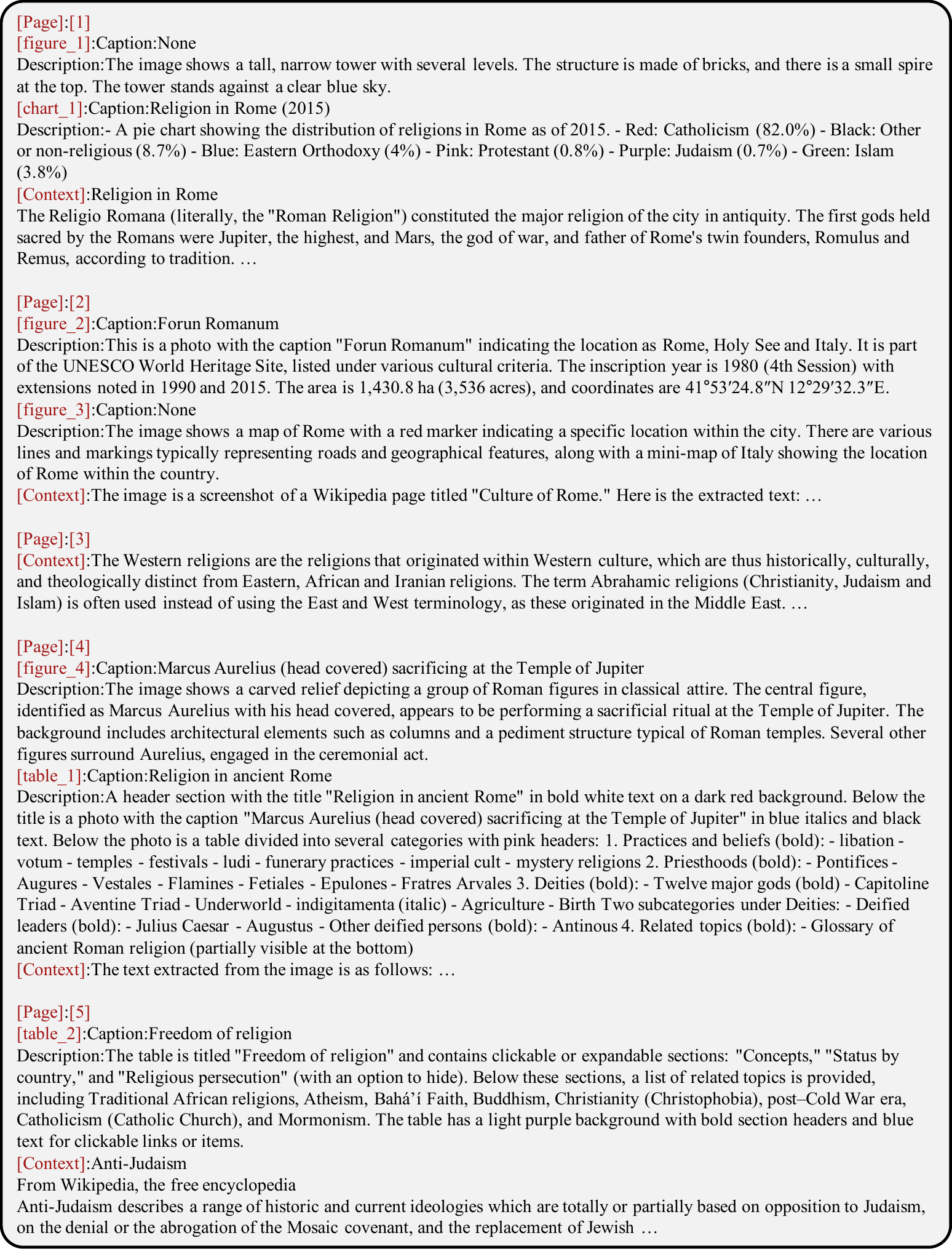}
    \caption{Input context example for Question and Answer generation.}
    \label{fig:input_context_ex}
\end{figure*}

We prompt Gemini 2.0 Flash Thinking with instructions focused on the following points for generating questions: 1) questions that target the core content of the given context; 2) questions whose answers can be derived from information distributed across multi pages; 3) when the context contains charts, tables, or figures, generate questions that integrate multiple modalities and contexts. We also include eight questions as few-shot examples in the prompt and direct the model to generate five questions for each context. Figure \ref{fig:input_context_ex} shows an example of input context, and Figure \ref{fig:prompt_gen_q} shows the prompt designed for the generation of questions. Subsequently, we sample three of the five questions generated for each cluster and use Gemini to verify and filter them, removing any questions that are ambiguous, not self-contained, or that referenced unseen context (e.g., “based on the document” or “according to the table”), and retain only those that passed this filtering step. The prompt for filtering questions is present in Figure \ref{fig:prompt_q_veri}. Out of the 90,000 generated train questions, 66,988 remain after verification, and out of the 6,000 test questions, 4,632 are filtered.

\subsection{Answer Generation}
For 80\% of the generated queries, we retrieve the top 5 pages using ColQwen to create answerable QA, while for the remaining 20\%, we retrieve the pages ranked 15th to 20th to construct challenging unanswerable QA pairs. We specifically select these lower-ranked pages because they contain partially relevant contexts, making the resulting unanswerable QA more difficult and realistic. These 5 retrieved pages form the context.

We perform answer generation using Gemini 2.0 Flash Thinking and Claude 3.7. The model is provided with a context and a single question, along with the following instructions in the prompt: 1) generate an answer by utilizing as much relevant information as possible from the given context in relation to the question; 2) when citing content from a specific page, include the page index (e.g., [1], [2]) as unique identifier within the response sentence; 3) when referencing charts, tables, or figures include the unique identifiers (e.g., [chart\_1], [table\_2], [figure\_3]) provided in the context within the response sentence; 4) structure the answer with an introduction, body, and conclusion, where the body is further divided into sections to provide a well-structured response format. For questions aimed at constructing unanswerable QA pairs, a different instruction is provided. While the rest of the process remains the same, the model is instructed to determine whether the question could be answered based solely on the given context, and to generate the reasoning behind this judgment to increase reliability. If the model determines the question to be answerable, it is instructed to generate an answer using the same instructions as for answerable QA pairs. We exclude the questions that are deemed answerable from this process. The prompts designed for the generation of answers are shown in Figure \ref{fig:prompt_gen_ans} and \ref{fig:prompt_gen_unans}.

\begin{figure*}[ht!]
    \centering
    \includegraphics[width=0.8\linewidth]{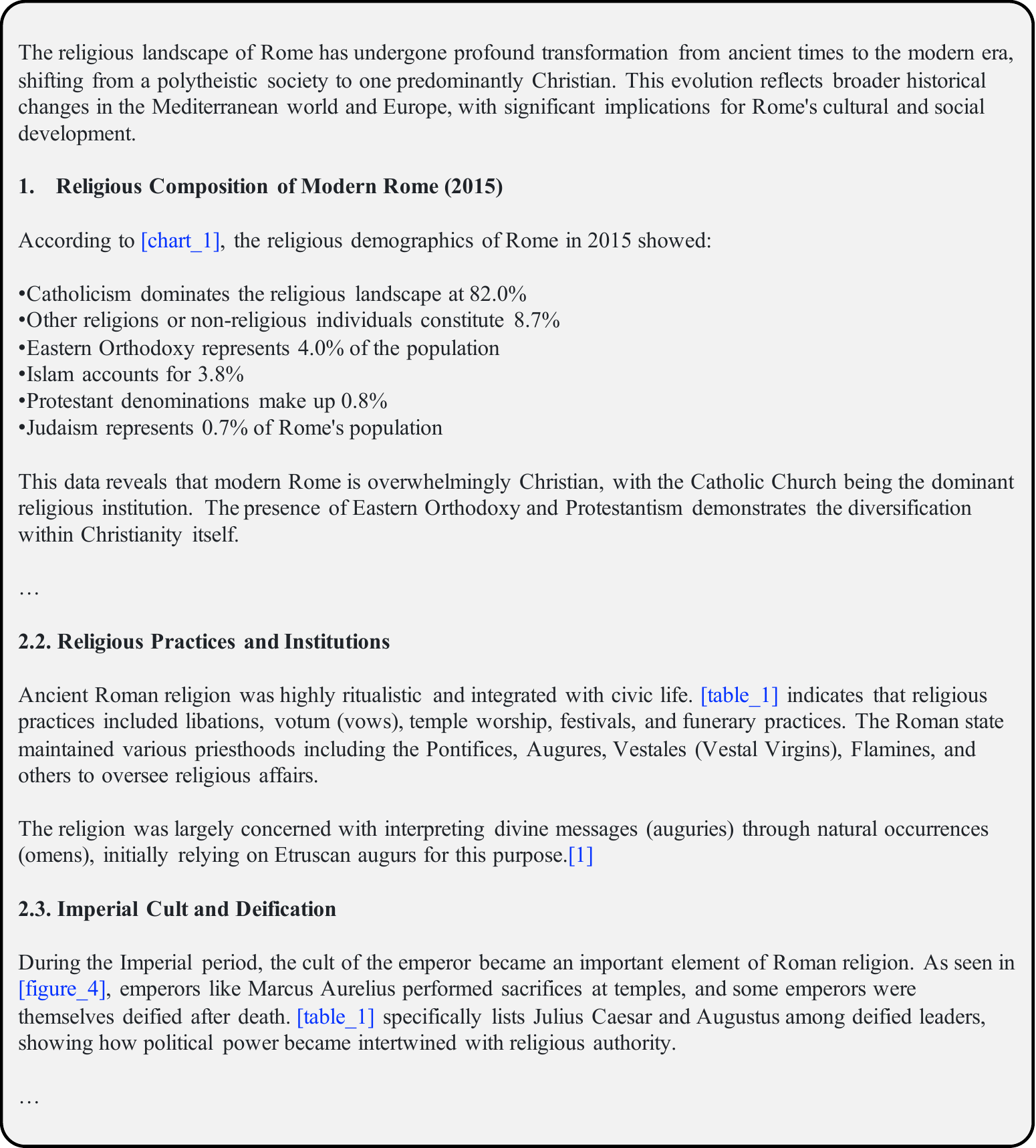}
    \caption{Answer example of VinQA. The \textcolor{blue}{blue} part indicates the citations of either the page numbers or the visual element identifiers of the charts, tables, and within the context.}
    \label{fig:answer_ex}
\end{figure*}

As a result, the train corpus contains 53,556 answerable and 10,554 unanswerable QA pairs, while the test corpus consists of 3,704 answerable QA and 751 unanswerable QA pairs. Figure \ref{fig:answer_ex} shows an example of the final answer generated from this process.

\subsection{Data Verification}
\label{sec:appendix_data_veri}


During the textual verification step, every Question–Context–Answer triple—where the context is in a textual form, as in the data generation process—is checked for 1) citation errors, 2) statements unsupported or contradicted by the context, 3) omissions of contextually relevant information, and 4) extraneous or query-irrelevant content; only data free of issues on all four criteria were preserved. For the training set, verification is performed solely by Gemini, whereas for the test set, verification is additionally conducted by Claude. Only data meeting all four criteria are retained. The prompt is shown in Figure \ref{fig:prompt_qa_veri_ans}.

During the visual verification step for the test set, whenever the supporting context included a chart, table, or figure, the corresponding image is supplied to the model as input. Visual verification is performed by both Gemini and Claude. The acceptance criteria are also adapted to the visual modality: the models must verify that 1) every image relevant to the question is properly used in the answer, and 2) each cited statement accurately reflects the information presented in its referenced image. Only QA pairs satisfying both visual-reasoning criteria are retained. The prompt is shown in Figure \ref{fig:prompt_vis_veri}.

For unanswerable QA data in the test set, we perform a distinct textual verification with Claude 3.7: each Question–Context pair is inspected to determine whether the context provides enough precise information to answer the question definitively. Any pair that met this condition is deemed incorrectly labeled and discarded. The prompt is shown in Figure \ref{fig:prompt_qa_veri_unans}.

After multi-step machine filtering, the training set comprised 39,700 answerable and 10,554 unanswerable QA pairs, while the test set comprised 1,822 answerable and 723 unanswerable pairs. To investigate the appropriate quantity of unanswerable data, we experiment with training configurations containing either 10K or 3K unanswerable QA pairs. Because using 10K led to a drop in M-GroSE scores, we reduced the number of unanswerable QA pairs to 3K in the final training setup. Additionally, text-reference-only QA pairs are sampled to balance the dataset. As a result, the training set comprises 39,700 answerable and 3,000 unanswerable QA pairs, and the test set comprises 1,312 answerable and 400 unanswerable QA pairs. 

For the test set, we apply an additional human verification stage on top of the textual and visual LLM-based checks described in the main paper. Four trained annotators manually inspect every QA pair following a structured guideline that checks (i) whether the bounding boxes of visual elements are correctly specified, (ii) whether visual citations are accurate and genuinely necessary to support the answer, and (iii) whether the supporting text is consistent with the cited visual elements and free of hallucinations. QA pairs that fail any of these checks are discarded. In total, 106 answerable and 1 unanswerable QA pairs are removed during this process. 





\section{Details of M-GroSE}
\label{sec:appendix_mgrose}
GroUSE~\cite{grouse} was originally designed to evaluate text-grounded answers, focusing solely on textual contexts. For our setting, we first refine the released GroUSE evaluation pipeline and prompt, as we found that some components led to suboptimal scoring behavior or were not directly suited to VinQA. Moreover, VinQA introduces a multimodal context where visual elements (e.g., tables, charts, figures) are interleaved within the retrieved pages, and the generated answers explicitly cite these visual elements. Therefore, we further adapt GroUSE’s evaluation methodology to this multimodal setting and refer to the resulting framework as Multimodal Grounded QA Scoring of Evaluators (M-GroSE).

Unlike the GroUSE pipeline, which uses an integrated scoring workflow, we directly evaluate all answerable QA instances along three criteria using an LLM-judge rubric based on GPT-5-mini:
(1) Relevancy — whether the answer appropriately addresses the question;
(2) Completeness — whether it incorporates all necessary information from the multimodal context;
and (3) Faithfulness — whether it remains consistent with the cited sources without hallucination.
We modify the evaluation prompt so that each criterion is assessed at the citation sentence level, enabling fine-grained judgments that consider every part of a long-form answer. The evaluation prompts for these three criteria are presented in Appendix~\ref{sec:appenidx_prompt_mgrose}.

GroUSE also includes a fourth criterion, Usefulness, which measures the helpfulness of an answer when the context is insufficient (i.e., adversarial or unanswerable settings). In VinQA, unanswerable questions are explicitly defined, and models are instructed not to provide additional speculative content. Therefore, Usefulness is not applicable. Instead, we separately evaluate the model’s behavior on unanswerable questions using the Unanswerability F1 metric, which measures whether the model correctly abstains when the provided context lacks sufficient evidence.

Finally, to extend the evaluation from text-only to multimodal grounding, we textualize both the textual and visual components of the context and adapt the LLM-judge prompts to evaluate not only textual reasoning but also the citations and explanations accompanying interleaved visual elements. This adaptation allows M-GroSE to jointly assess the groundedness of answers in multimodal document QA.

\section{Prompt Template}
\label{sec:appendix_prompt}

\subsection{Prompts for Data generation}
\label{sec:appenidx_prompt_data_const}
\begin{figure*}[ht!]
    \centering
    \includegraphics[width=0.8\linewidth]
    {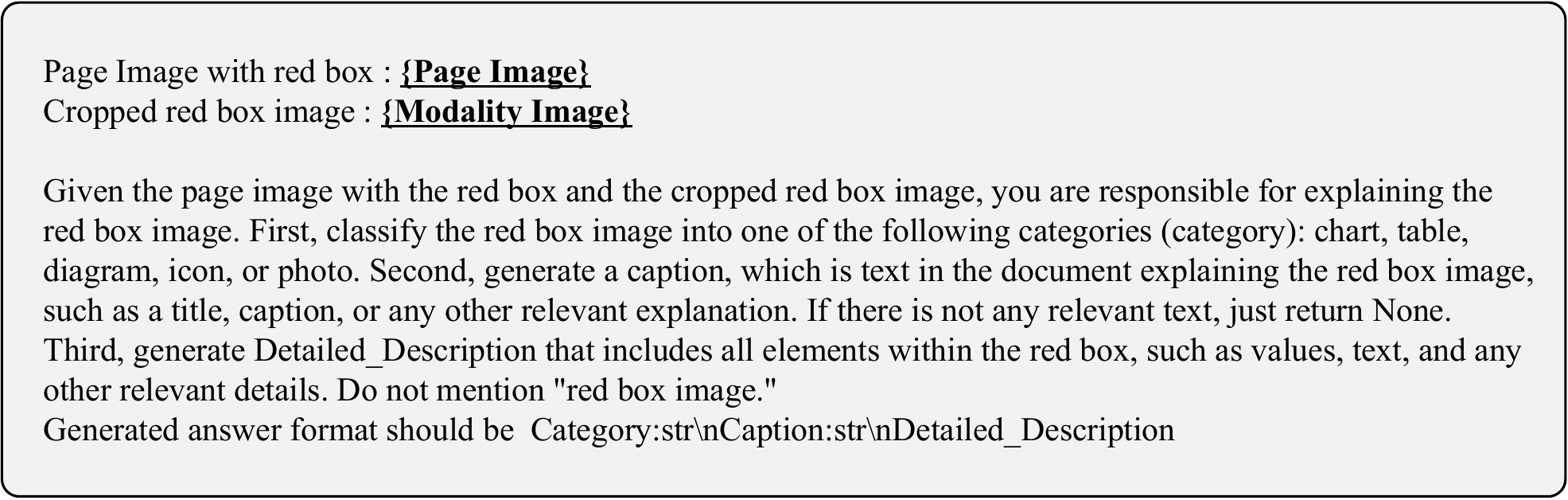}
    \caption{The prompt for generating visual-element's class label, caption, and description.}
    \label{fig:modal_desc_cap_prompt}
\end{figure*}

\begin{figure*}[ht!]
    \centering
    \includegraphics[width=0.8\linewidth]{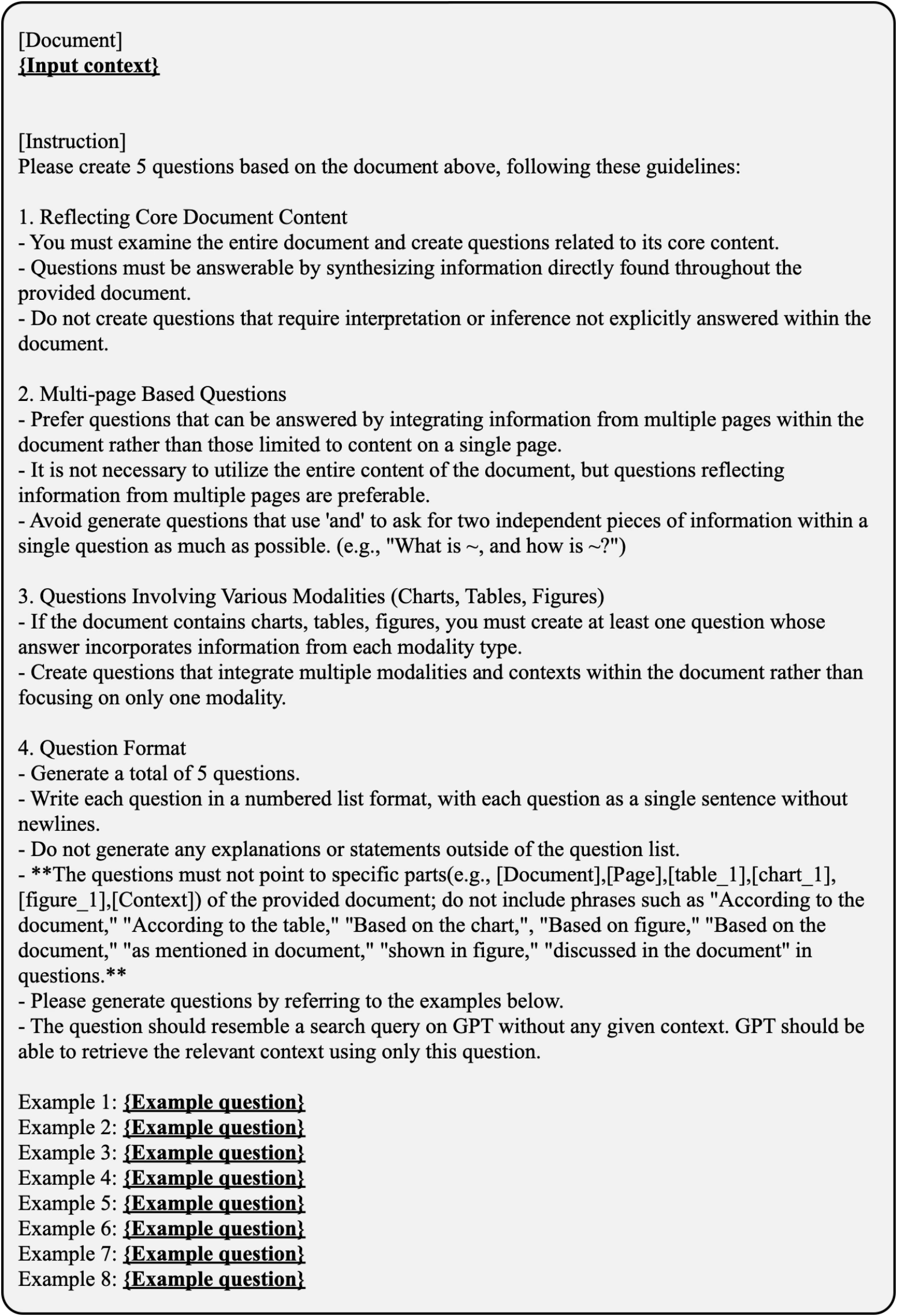}
    \caption{The prompt for Question generation. The text marked with both bold and underline represents the parts provided as prompt inputs.}
    \label{fig:prompt_gen_q}
\end{figure*}

\begin{figure*}[ht!]
    \centering
    \includegraphics[width=0.8\linewidth]{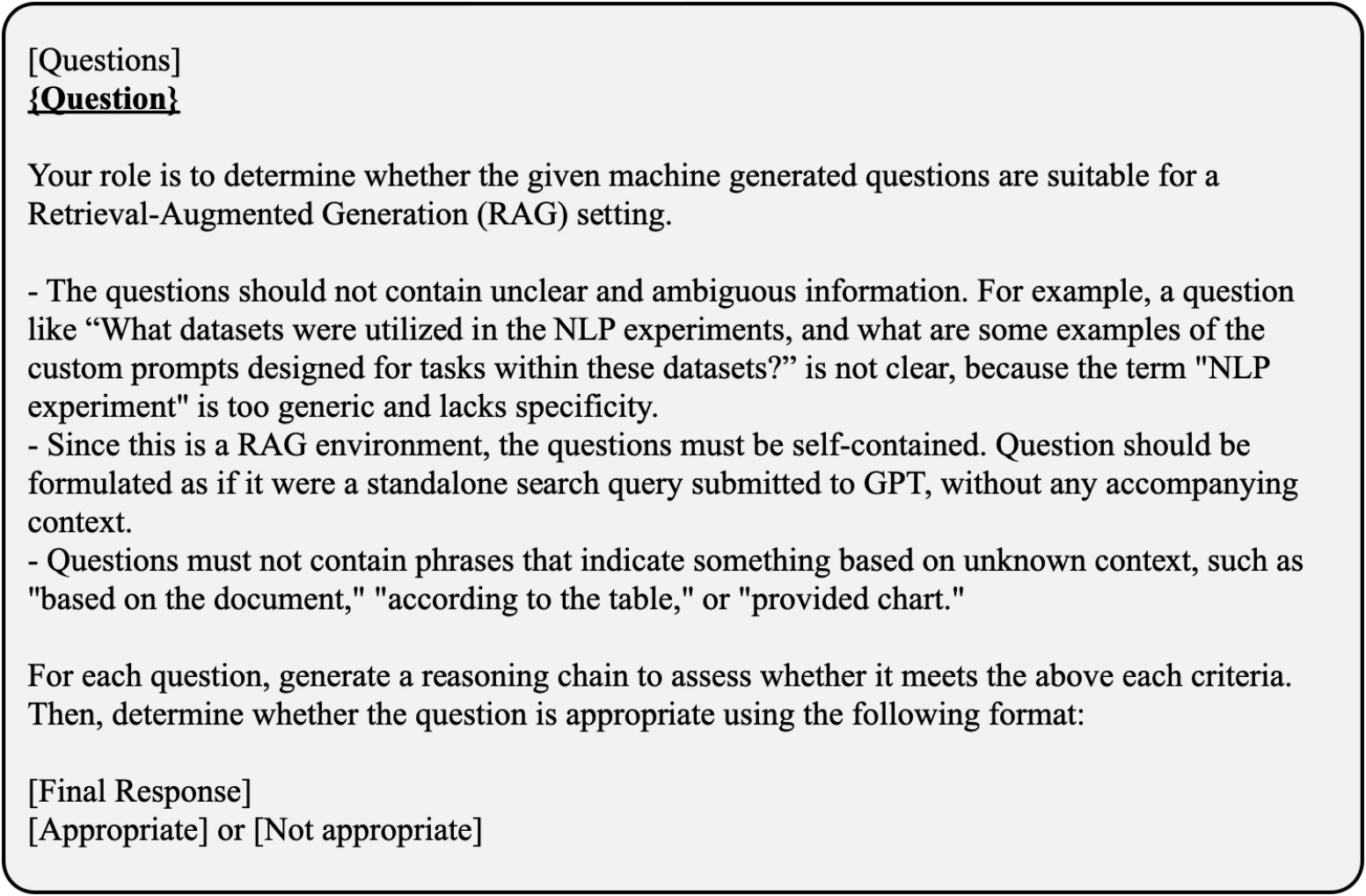}
    \caption{The prompt for Question filtering.}
    \label{fig:prompt_q_veri}
\end{figure*}

\begin{figure*}[ht!]
    \centering
    \includegraphics[width=0.8\linewidth]{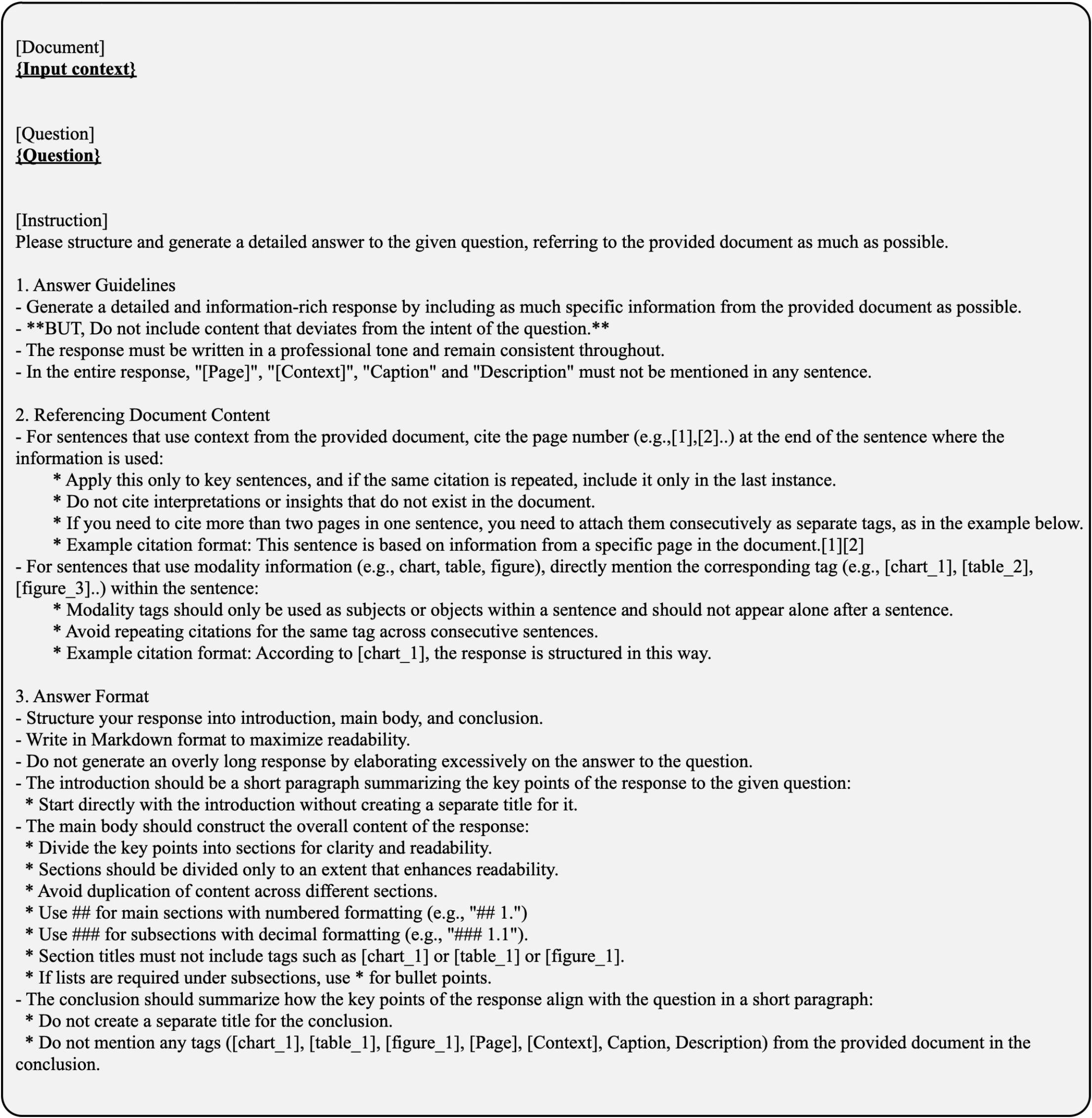}
    \caption{The prompt for Answer generation (Answerable QA).}
    \label{fig:prompt_gen_ans}
\end{figure*}

\begin{figure*}[ht!]
    \centering
    \includegraphics[width=0.8\linewidth]{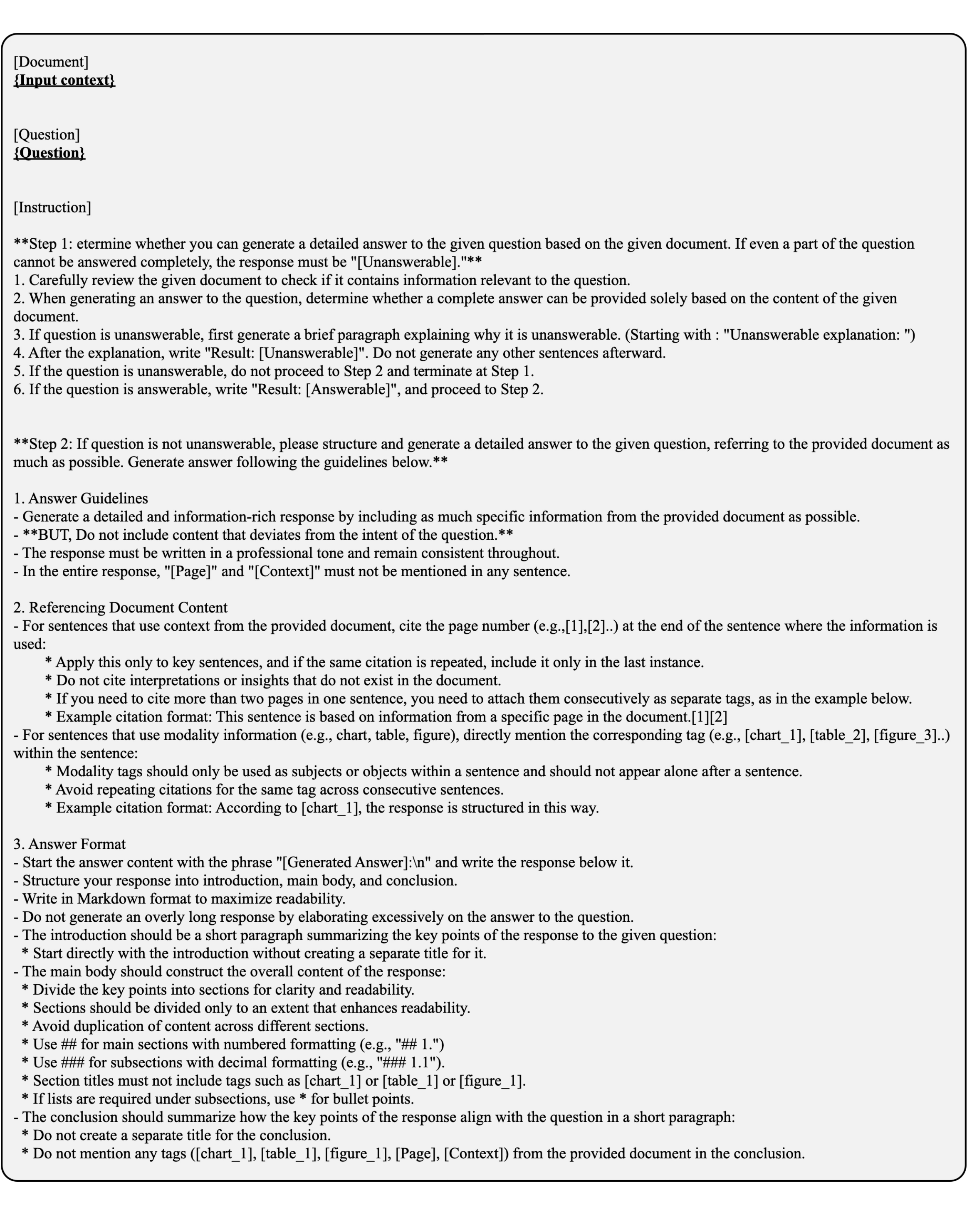}
    \caption{The prompt for Answer generation (Unanswerable QA).}
    \label{fig:prompt_gen_unans}
\end{figure*}

Figures \ref{fig:modal_desc_cap_prompt}--\ref{fig:prompt_gen_unans} show the prompts used during the VinQA dataset generation process.

\subsection{Prompts for Data verification}
\label{sec:appenidx_prompt_data_verifi}
\begin{figure*}[ht!]
    \centering
    \includegraphics[width=0.8\linewidth]{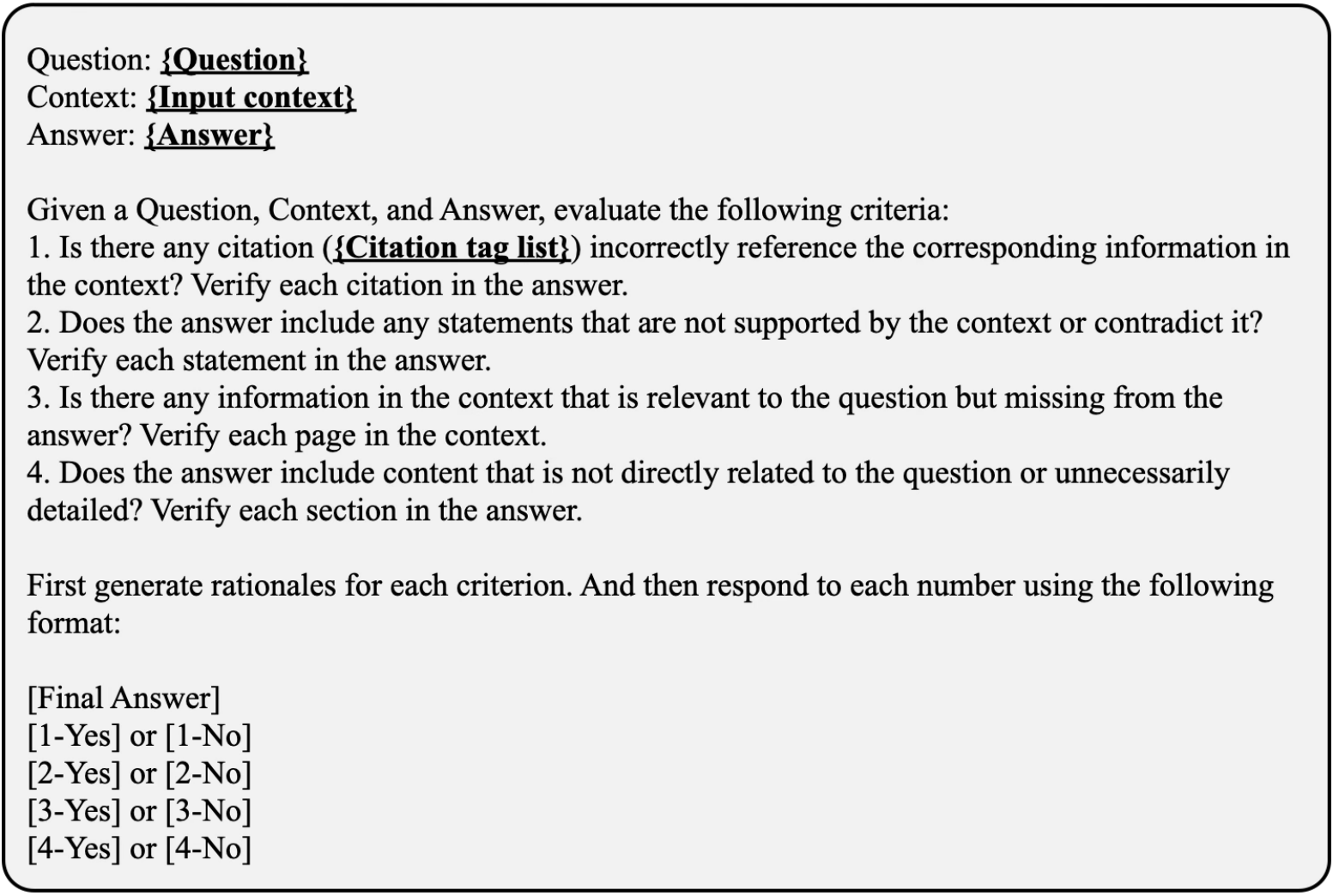}
    \caption{The prompt for Textual verification (Answerable QA). Citation tag list indicates all page number tags and the chart, table, and figure modality tags that appear in the answer.}
    \label{fig:prompt_qa_veri_ans}
\end{figure*}

\begin{figure*}[ht!]
    \centering
    \includegraphics[width=0.8\linewidth]{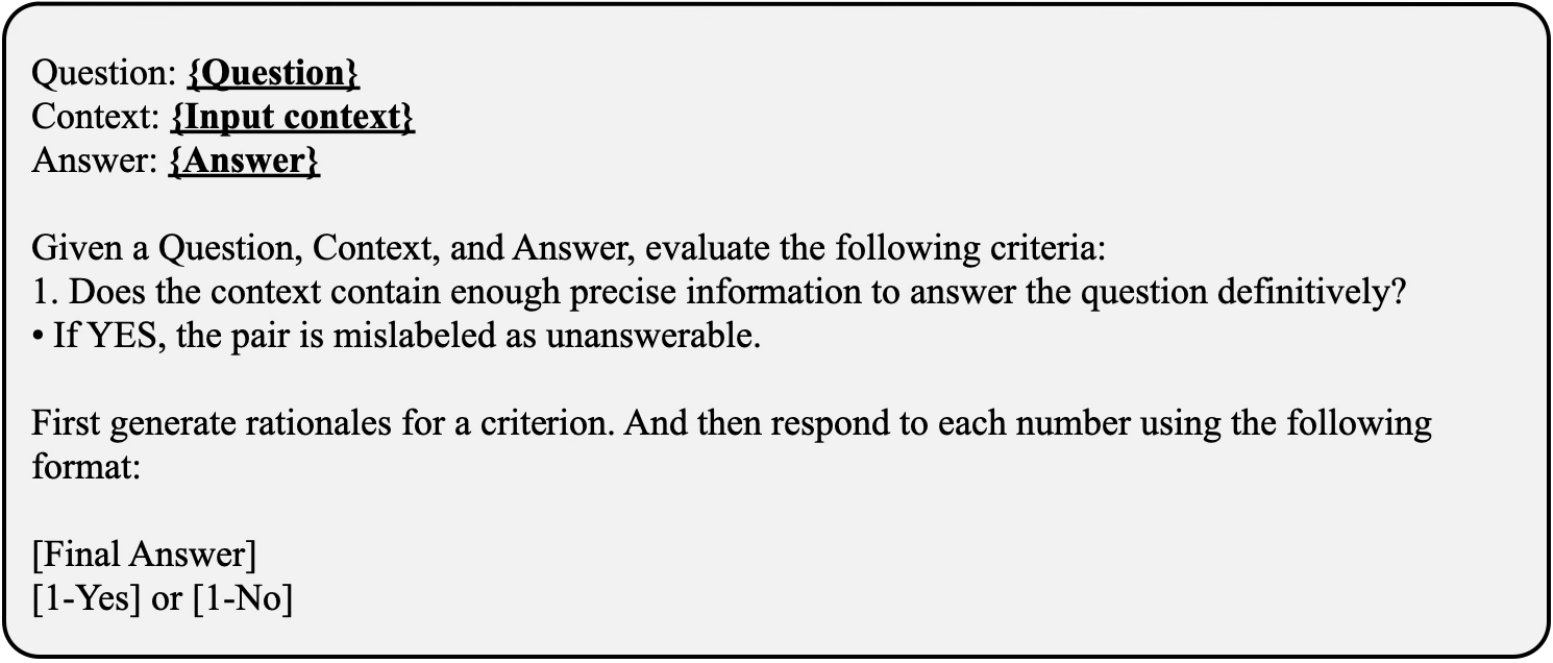}
    \caption{The prompt for Textual verification (Unanswerable QA).}
    \label{fig:prompt_qa_veri_unans}
\end{figure*}

\begin{figure*}[ht!]
    \centering
    \includegraphics[width=0.8\linewidth]{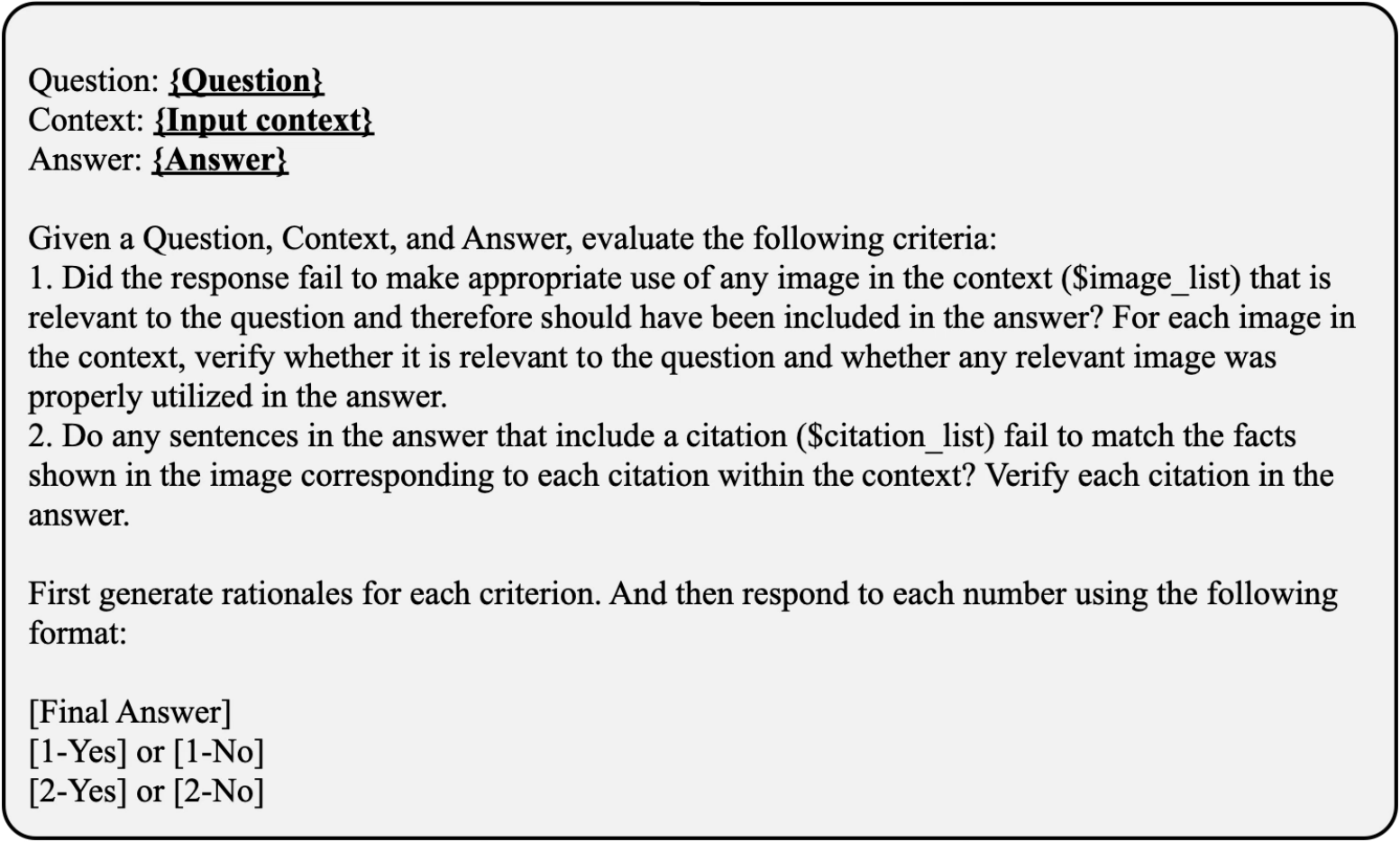}
    \caption{The prompt for Visual verification. For visual verification, the input context differs from textual verification: any chart, figure, or table modality found in the context is replaced with its image, so the resulting context contains visual elements interleaved with the text.}
    \label{fig:prompt_vis_veri}
\end{figure*}

Figures \ref{fig:prompt_qa_veri_ans}, \ref{fig:prompt_qa_veri_unans}, and \ref{fig:prompt_vis_veri} show the prompts used for data verification.

\subsection{Prompt examples for encoding method}
\label{sec:appenidx_prompt_ex_enc}

\begin{figure*}[ht!]
    \centering
    \includegraphics[width=0.8\linewidth]{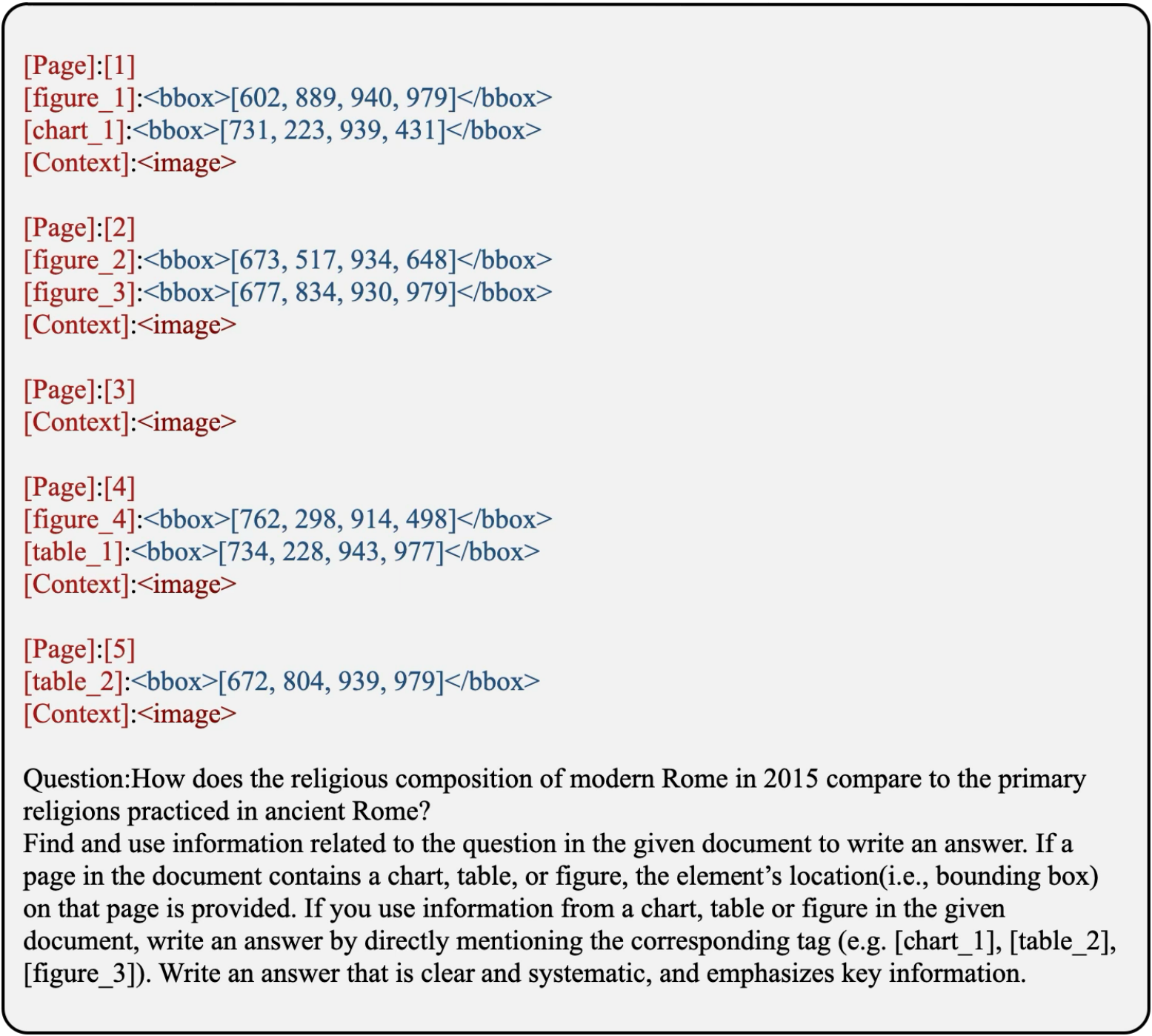}
    \caption{The prompt example for \textit{Page Encoding} method. The $<$image$>$ part refers to the corresponding page image input that is converted into visual tokens.}
    \label{fig:page_encoding_train_prompt}
\end{figure*}

\begin{figure*}[ht!]
    \centering
    \includegraphics[width=0.8\linewidth]{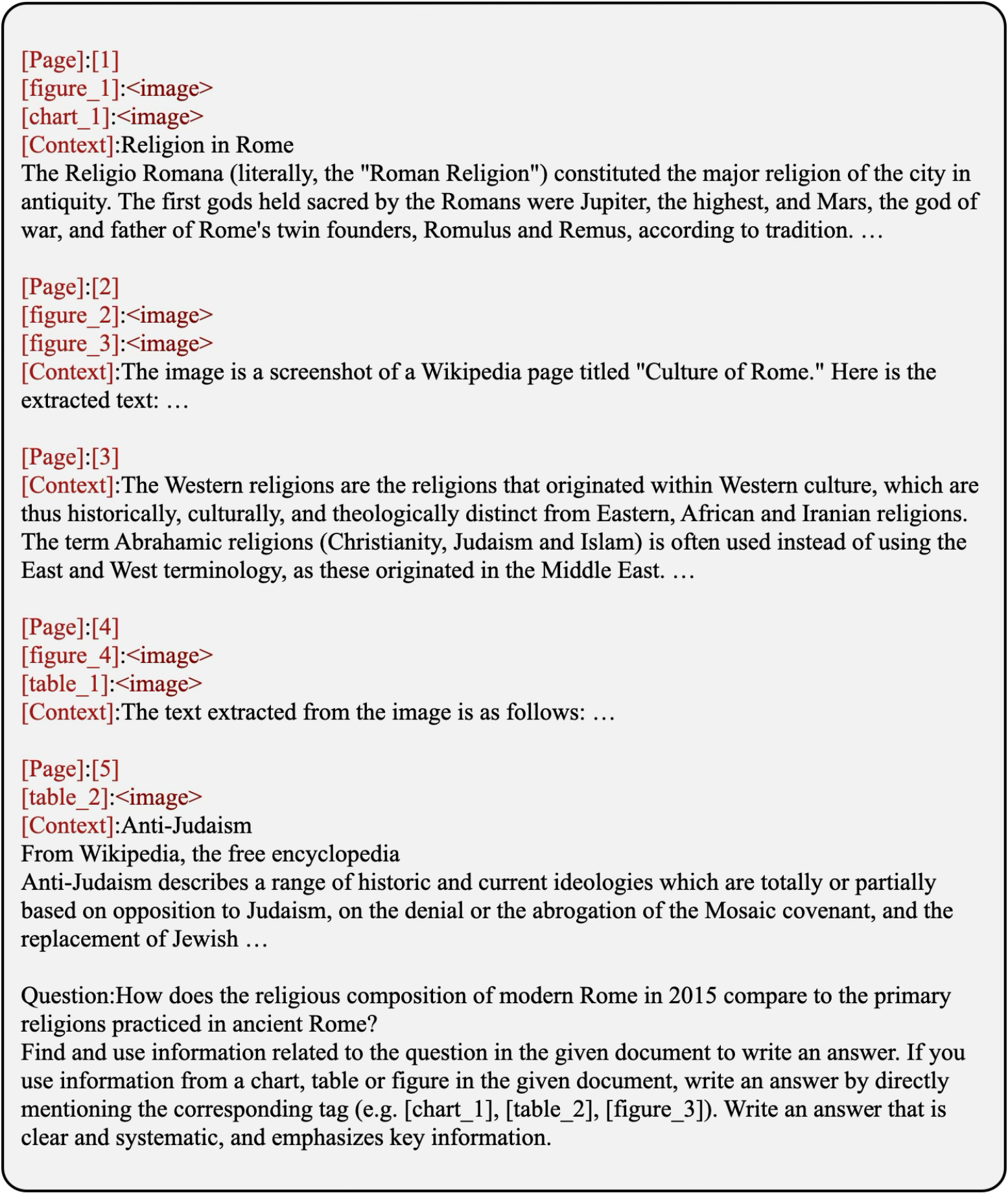}
    \caption{The prompt example for \textit{Modality Encoding} method. The $<$image$>$ part refers to the input corresponding to each visual element identifier, which is converted into visual tokens.} 
    \label{fig:modality_encoding_train_prompt}
\end{figure*}

Figure \ref{fig:page_encoding_train_prompt} shows an example prompt input for the \textit{Page Encoding} method, while Figure \ref{fig:modality_encoding_train_prompt} corresponds to the \textit{Modality Encoding} method.

\subsection{Prompts for M-GroSE}
\begin{figure*}[ht!]
    \centering
    \includegraphics[width=0.8\linewidth]
    {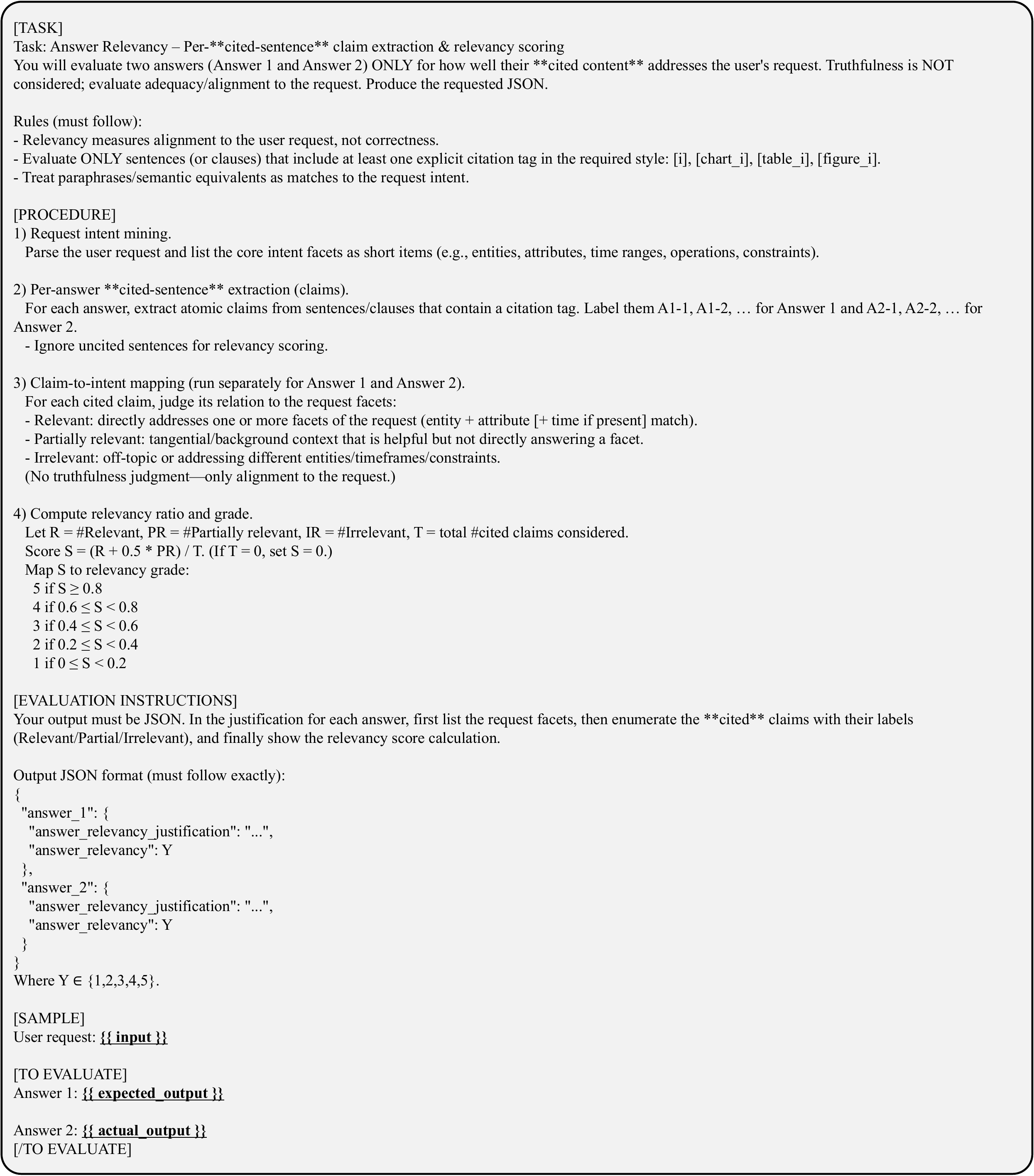}
    \caption{The prompt used for evaluating relevancy in M-GroSE.}
    \label{fig:prompt_mgrose_relevancy}
\end{figure*}

\begin{figure*}[ht!]
    \centering
    \includegraphics[width=0.8\linewidth]
    {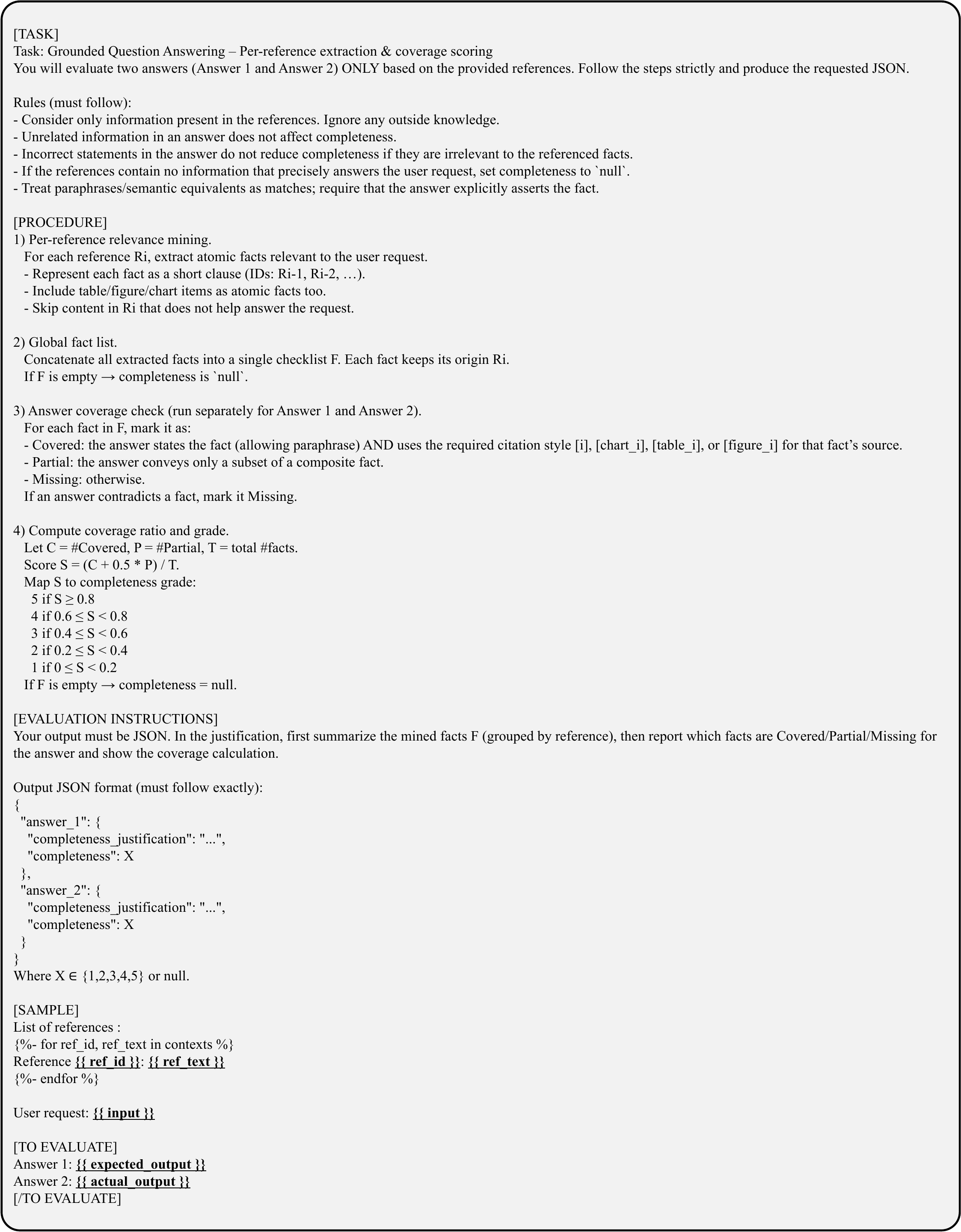}
    \caption{The prompt used for evaluating completeness in M-GroSE.}
    \label{fig:prompt_mgrose_completeness}
\end{figure*}

\begin{figure*}[ht!]
    \centering
    \includegraphics[width=0.8\linewidth]
    {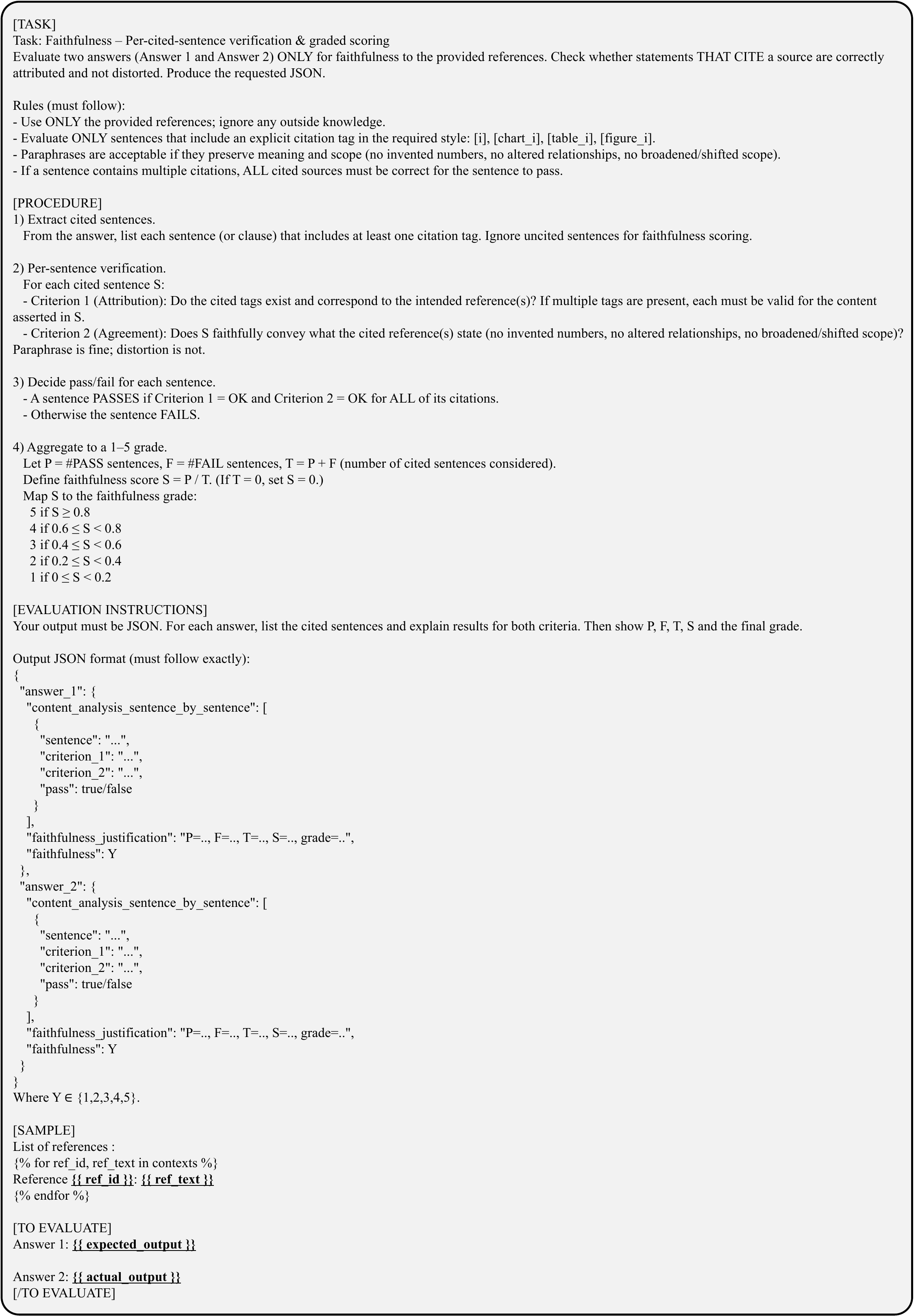}
    \caption{The prompt used for evaluating answer faithfulness in M-GroSE.}
    \label{fig:prompt_mgrose_faithfulness}
\end{figure*}

Figures \label{sec:appenidx_prompt_mgrose}
 \ref{fig:prompt_mgrose_relevancy}, \ref{fig:prompt_mgrose_completeness}, and \ref{fig:prompt_mgrose_faithfulness} show the prompts used for M-GroSE.

\subsection{Prompts for Visual G-Eval}
\label{sec:appenidx_prompt_eval}
\begin{figure*}[ht!]
    \centering
    \includegraphics[width=0.8\linewidth]
    {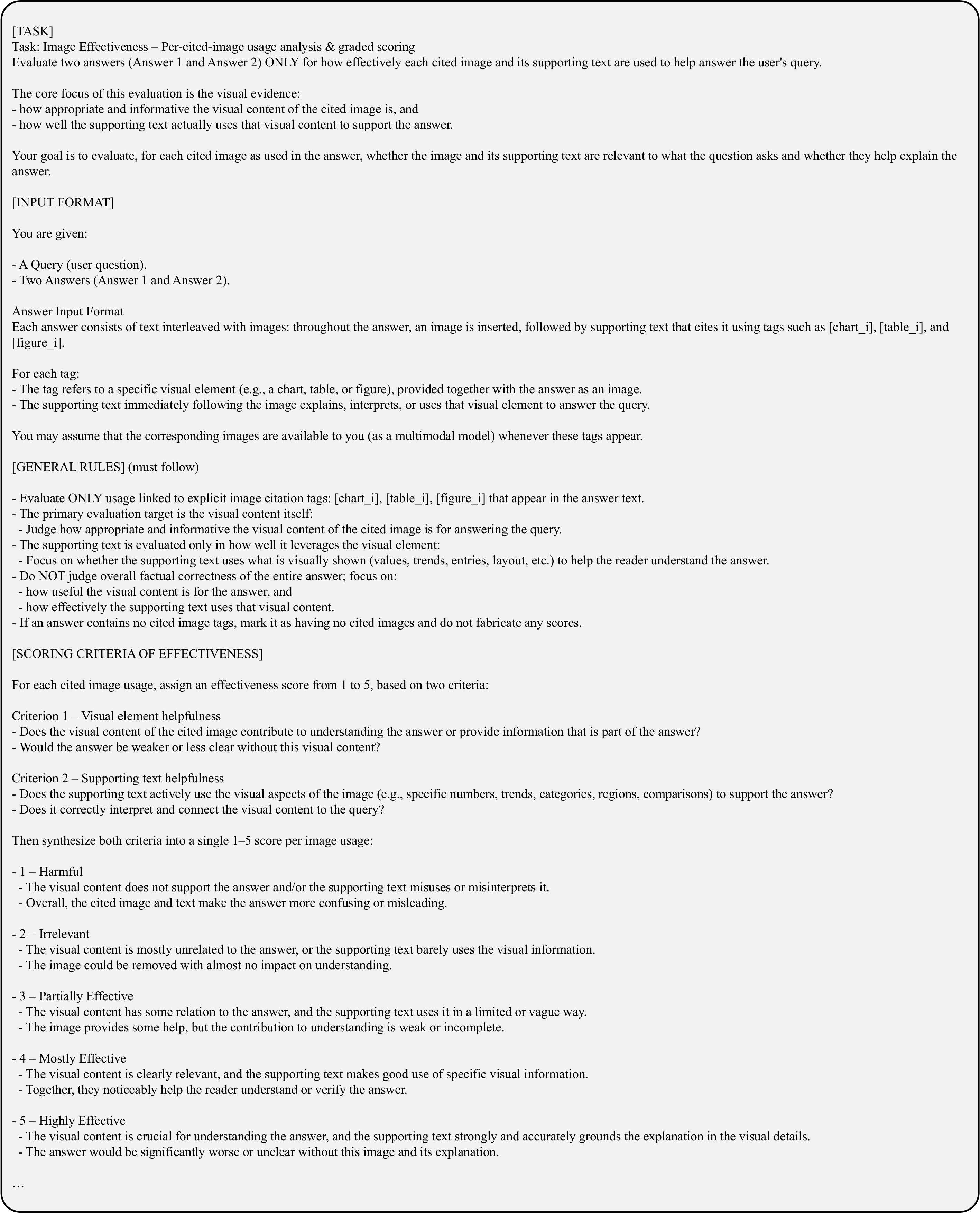}
    \caption{The Visual G-Eval prompt to assess Effectiveness. (Part 1)}
    \label{fig:img_citation_Effectiveness1}
\end{figure*}

\begin{figure*}[ht!]
    \centering
    \includegraphics[width=0.8\linewidth]
    {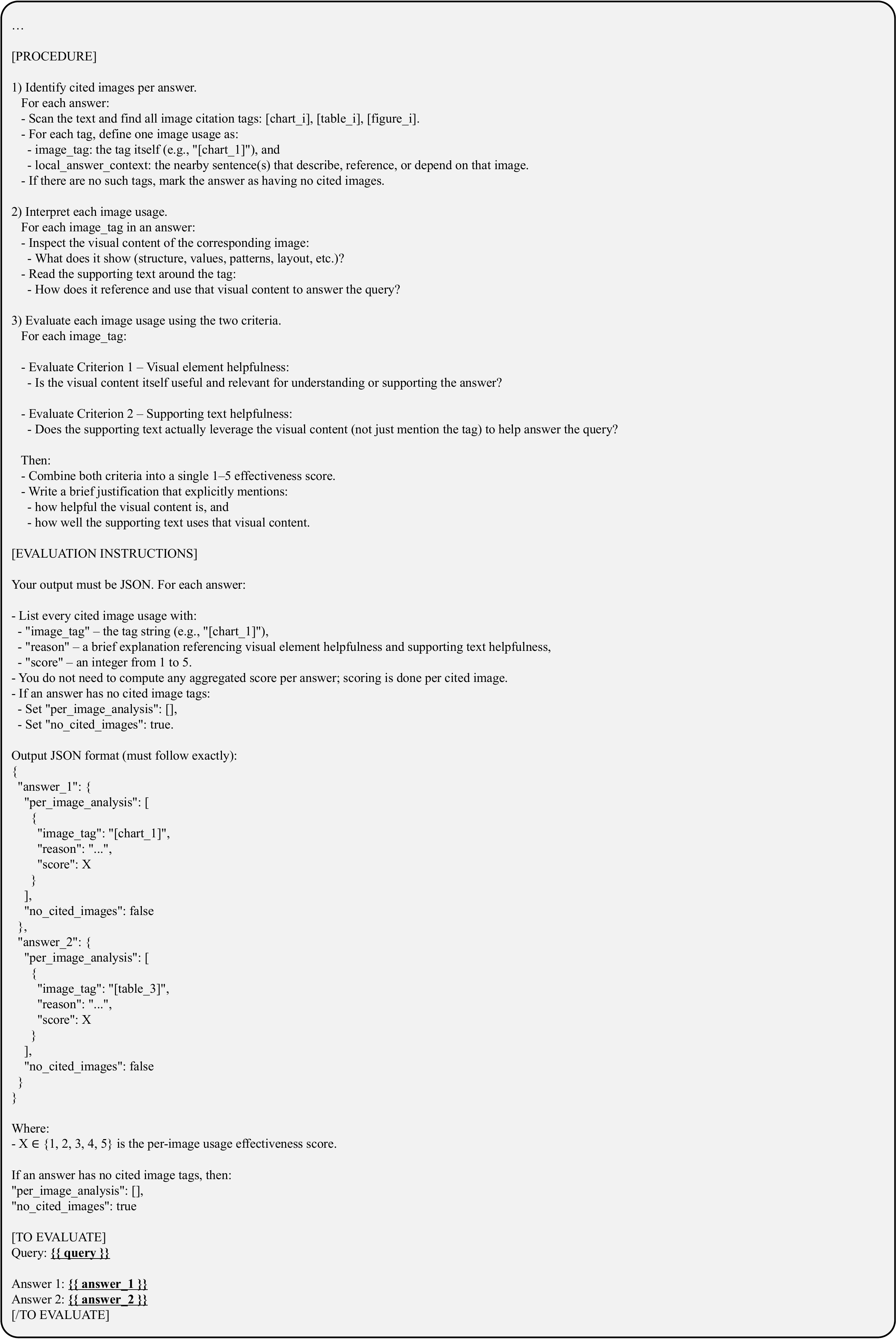}
    \caption{The Visual G-Eval prompt to assess Effectiveness (Part 2).}
    \label{fig:img_citation_Effectiveness2}
\end{figure*}

\begin{figure*}[ht!]
    \centering
    \includegraphics[width=0.8\linewidth]
    {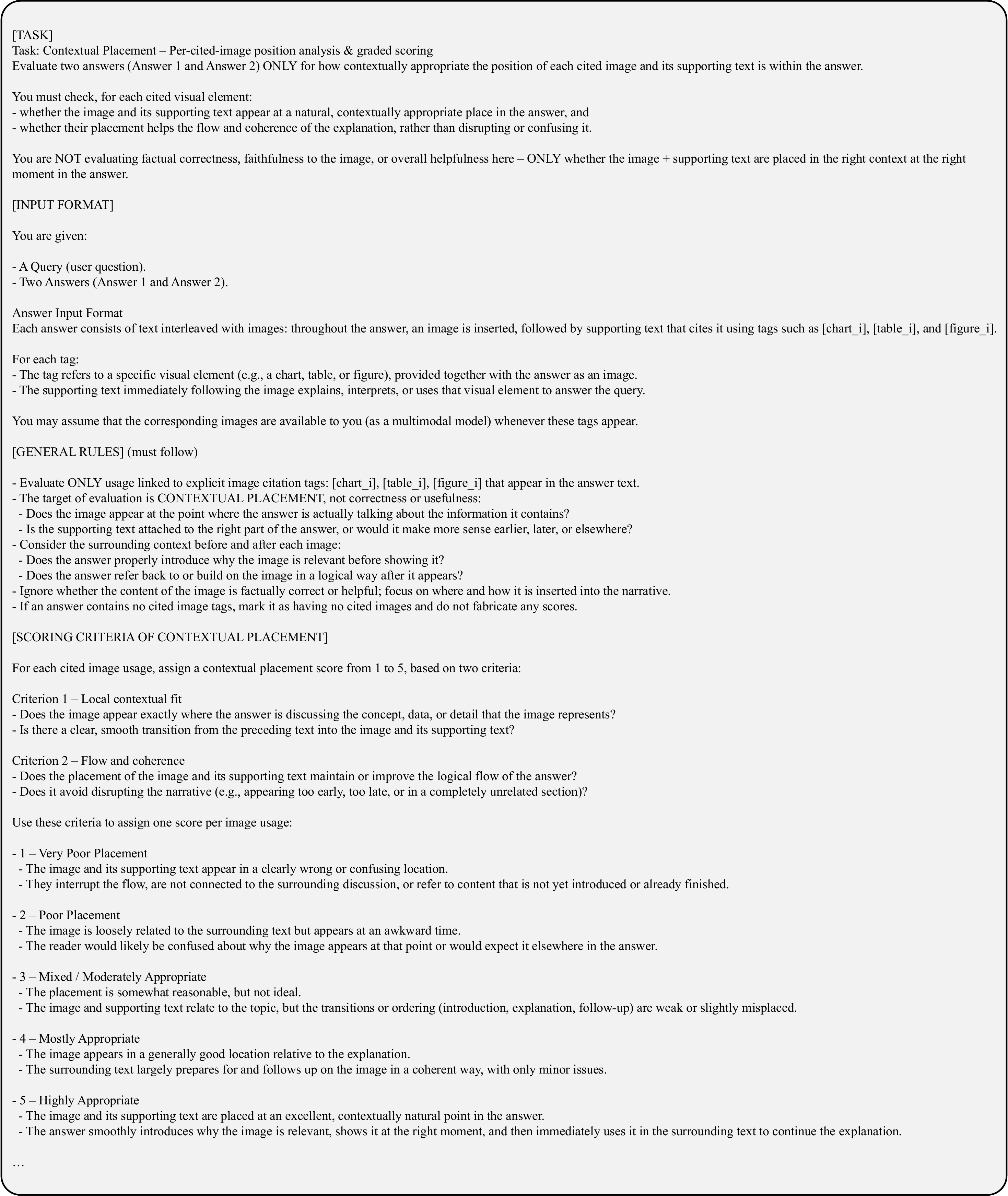}
    \caption{The Visual G-Eval prompt to assess Position (Part 1).}
    \label{fig:img_citation_Position_Correctness1}
\end{figure*}

\begin{figure*}[ht!]
    \centering
    \includegraphics[width=0.8\linewidth]
    {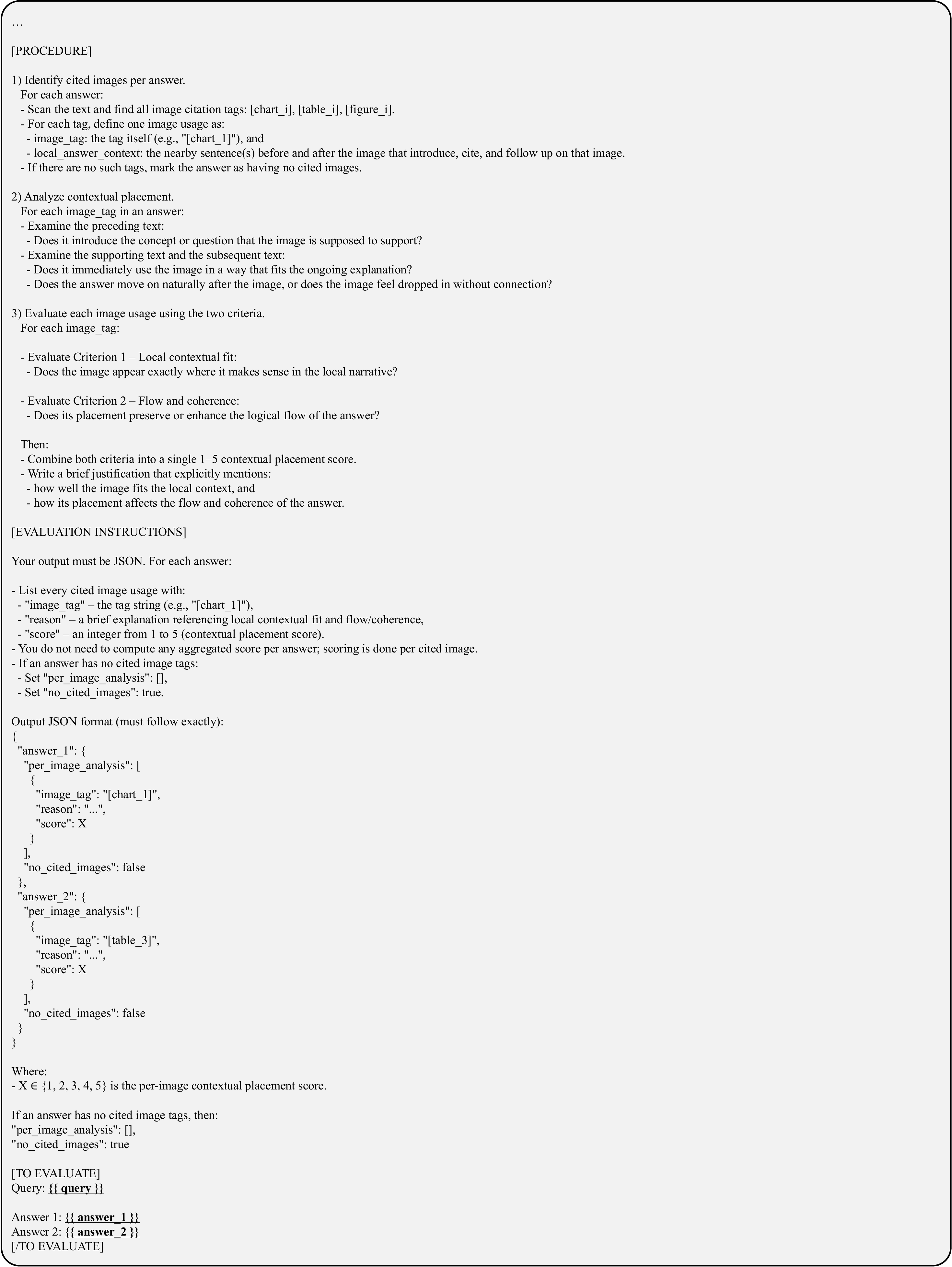}
    \caption{The Visual G-Eval prompt to assess Position (Part 2).}
    \label{fig:img_citation_Position_Correctness2}
\end{figure*}

\begin{figure*}[ht!]
    \centering
    \includegraphics[width=0.8\linewidth]
    {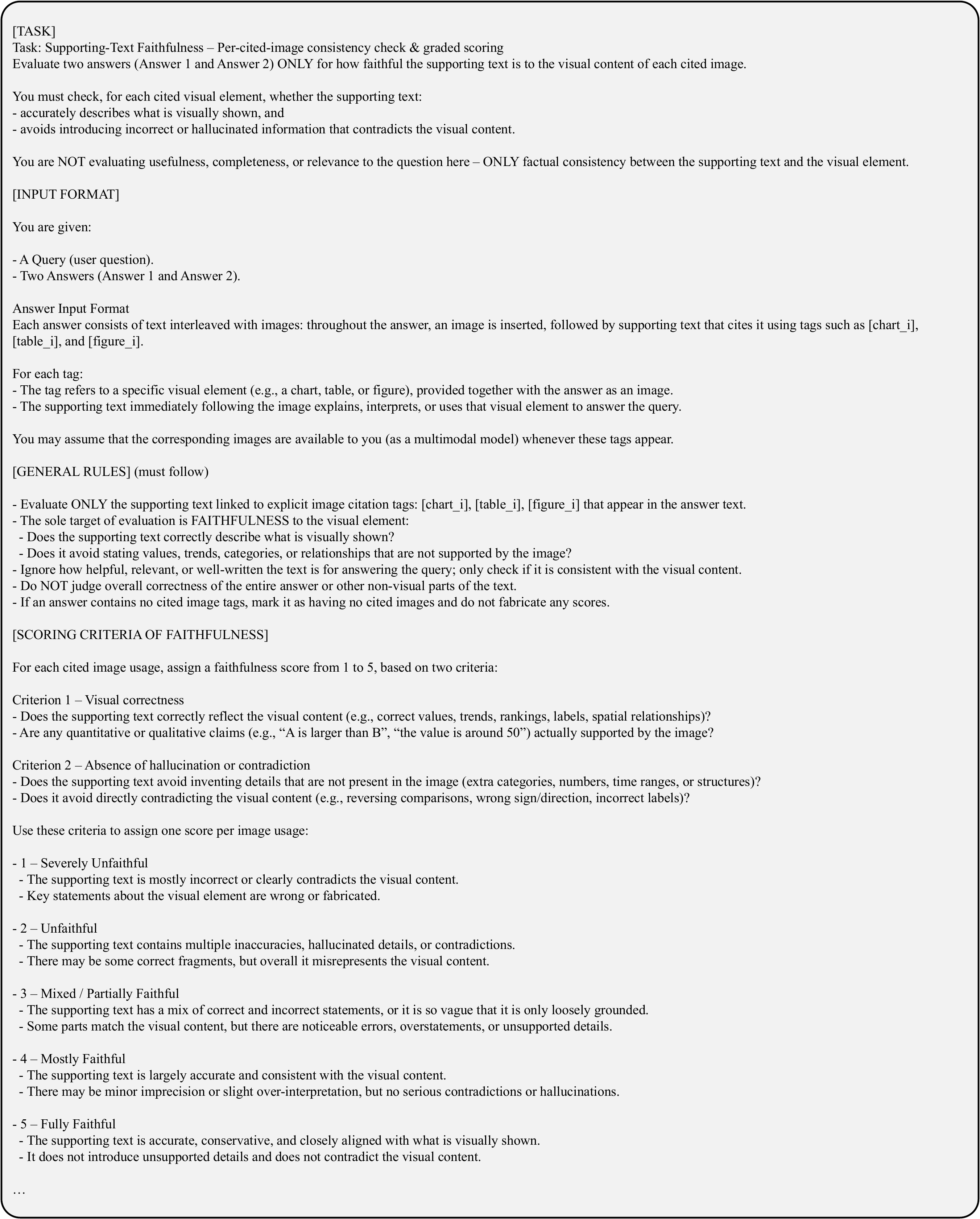}
    \caption{The Visual G-Eval prompt to assess Faithfulness (Part 1).}
    \label{fig:img_citation_Expression_faithfulness1}
\end{figure*}

\begin{figure*}[ht!]
    \centering
    \includegraphics[width=0.8\linewidth]
    {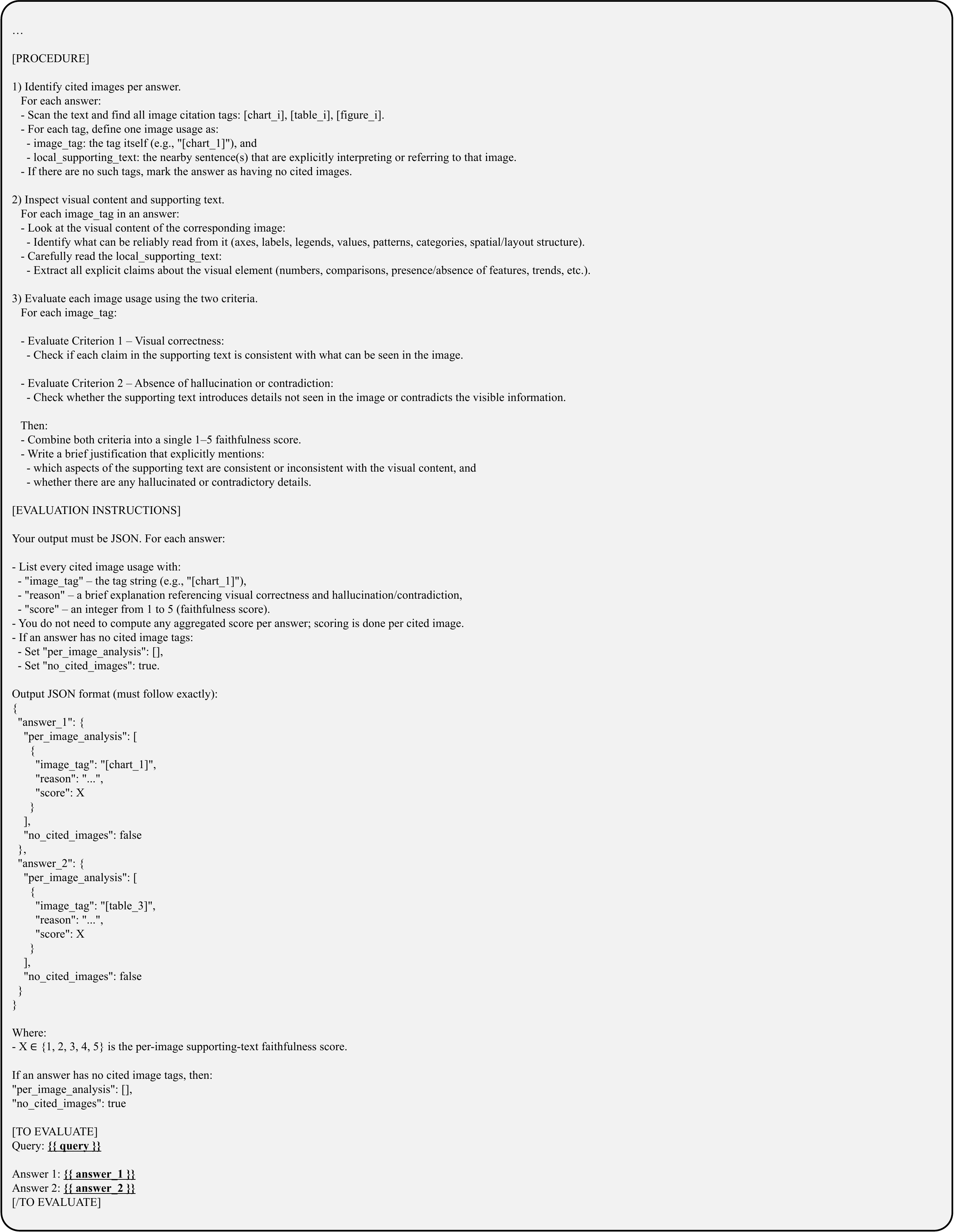}
    \caption{The Visual G-Eval prompt to assess Faithfulness (Part 2).}
    \label{fig:img_citation_Expression_faithfulness2}
\end{figure*}

Figures \ref{fig:img_citation_Effectiveness1}--\ref{fig:img_citation_Expression_faithfulness2} show the prompts used for Visual G-Eval.

\clearpage

\section{Correlation between Visual G-Eval and Human Judgments}
M-GroSE extends human-aligned GroUSE~\cite{grouse} to multimodal LLM-based judging by textualizing visual elements. However, since Visual G-Eval lacks comparable validation, we conduct a targeted human study on 50 samples. Two human judges independently rate \textit{effectiveness}, \textit{position}, and \textit{faithfulness}, and we measure rank correlation between Visual G-Eval and human scores. As shown in \cref{tab:geval_human_corr}, Visual G-Eval correlates well with human judgments (Spearman $\rho$ = 0.68--0.80), while human inter-annotator agreement is moderate to high (QW $\kappa$ = 0.53--0.87).

\begin{table}[t]
\centering
\small
\resizebox{0.5\columnwidth}{!}{
\begin{tabular}{lcc}
\toprule
\textbf{Dimension} & \textbf{Spearman $\rho$} & \textbf{QW Cohen's $\kappa$} \\
\midrule
Effectiveness & 0.7962 & 0.5333 \\
Position      & 0.6775 & 0.6853 \\
Faithfulness  & 0.7126 & 0.8676 \\
\bottomrule
\end{tabular}
}
\vspace{-2mm}
\caption{Correlation between Visual G-Eval and human judgments on 50 samples, and inter-annotator agreement (quadratic-weighted Cohen's $\kappa$) between two human judges.}
\label{tab:geval_human_corr}
\vspace{-5mm}
\end{table}

\section{Qualitative Analysis}
\label{sec:qualitative_analysis}

Figures \ref{fig:success_case}--\ref{fig:err_case2} provide the output examples of our model, including comparison with baselines and failure cases.  

\subsection{Evaluation}
\begin{figure*}[ht!]
    \centering
    \includegraphics[width=1.0\linewidth]
    {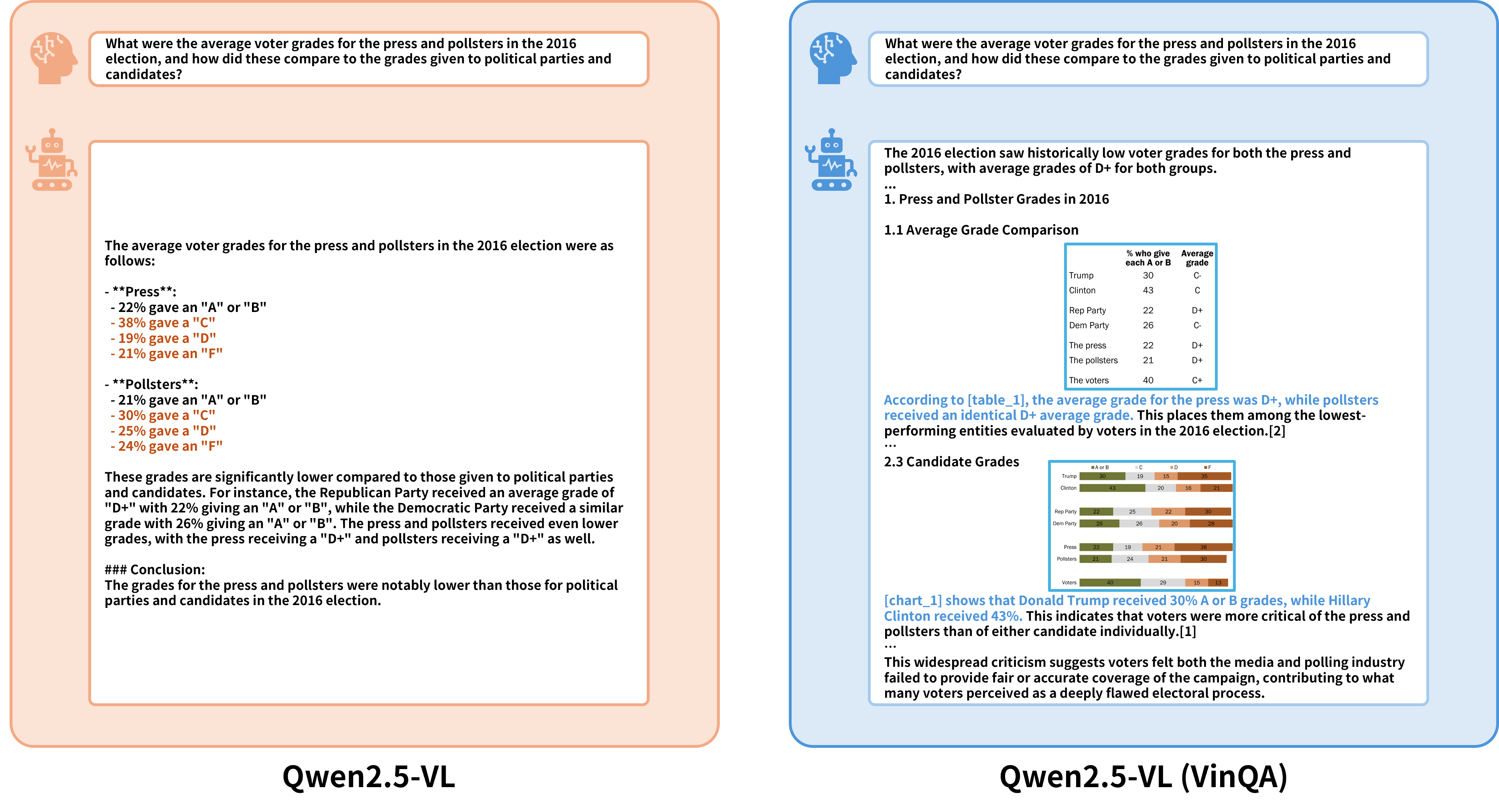}
    \caption{Comparison of predicted answers between Qwen2.5-VL-7B and our model trained on VinQA. The \textcolor{blue}{blue} and \textcolor{red}{red} text respectively indicate the portions correctly and incorrectly predicted by the model.}
    \label{fig:success_case}
\end{figure*}

\begin{figure*}[ht!]
    \centering
    \includegraphics[width=1.0\linewidth]
    {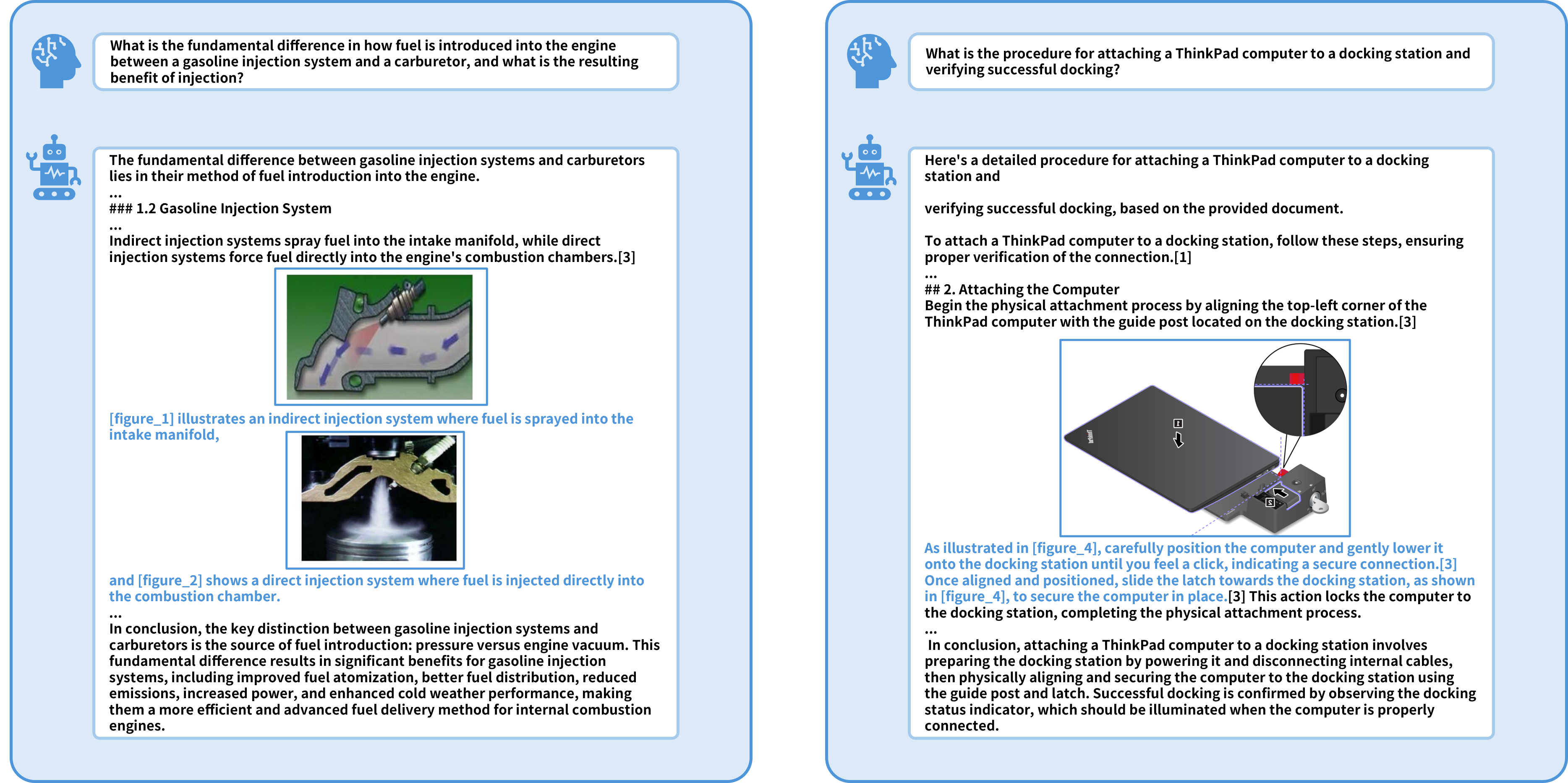}
    \caption{Inference samples of our model trained with VinQA.}
    \label{fig:example_integrated}
\end{figure*}

Figure~\ref{fig:success_case} presents evaluation results from Qwen2.5-VL-7B and our model trained on VinQA. Compared to Qwen2.5-VL-7B, our model more effectively cites visual elements and generates descriptions in appropriate positions, resulting in more accurate and information-rich responses. Figure~\ref{fig:example_integrated} shows successful results of our model. Specifically, our model improved readability by effectively citing user manual-related images and diagrams that are difficult to clearly comprehend through textual information alone.

\subsection{Error Analysis}

\begin{figure*}[ht!]
    \centering
    \includegraphics[width=0.8\linewidth]
    {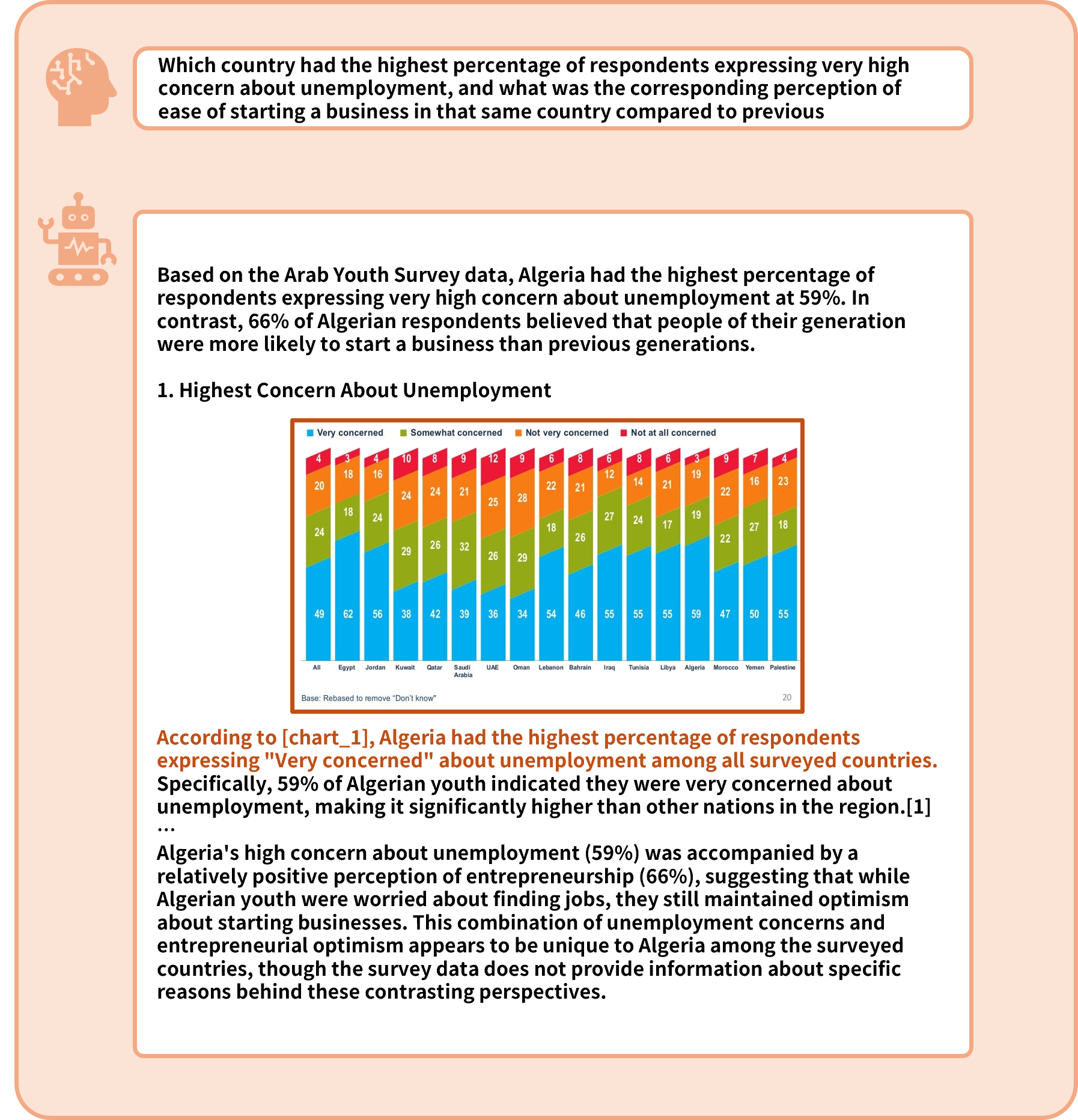}
    \caption{Faithfulness error in the predicted answer from our model trained on VinQA. The text highlighted in \textcolor{red}{red} inaccurately describes the corresponding visual element.}
    \label{fig:err_case1}
\end{figure*}

\begin{figure*}[ht!]
    \centering
    \includegraphics[width=1.0\linewidth]
    {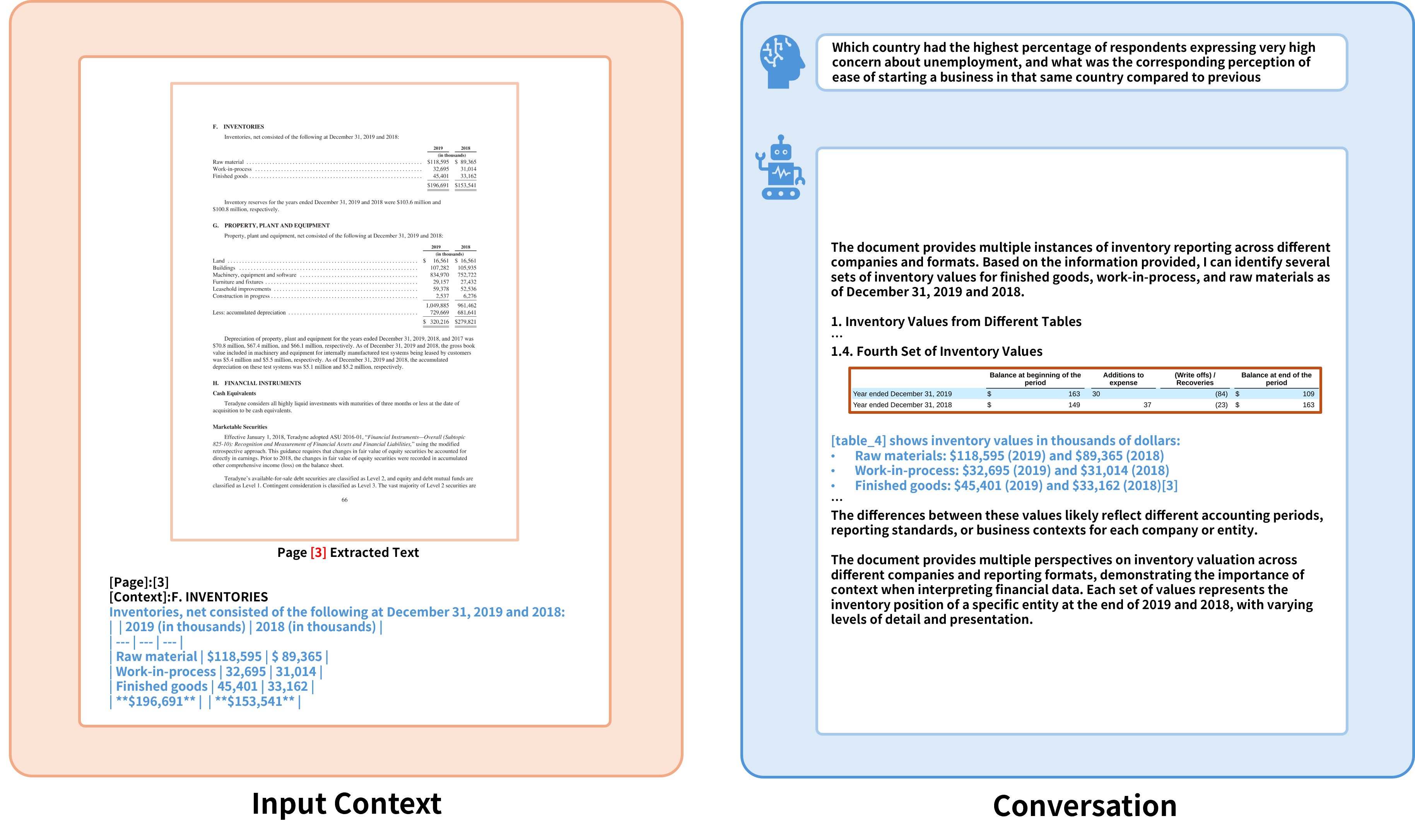}
    \caption{Citation error in the predicted answer from our model trained on VinQA. The model correctly retrieves question-relevant content from Page~[3], as shown in the \textcolor{blue}{blue} text, and uses it appropriately in the answer. However, it incorrectly cites an unrelated visual element, highlighted in \textcolor{red}{red}.}
    \label{fig:err_case2}
\end{figure*}

To identify the challenges, we analyze the inference results of our model. Representative failure cases are presented in Figure \ref{fig:err_case1} and \ref{fig:err_case2}.

\textbf{Faithfulness}
Our model generally shows strong performance in citing relevant visual elements and providing appropriate explanations. However, in some examples, hallucinations related to visual elements are observed. In Figure \ref{fig:err_case1}, our model successfully retrieved the visual element relevant to the given question but failed to recognize its detailed values. Such issues primarily occur when the input image exhibits a complex structure, such as charts containing a large amount of information in small text.

\textbf{Citation}
While our model generally performs well, it occasionally cites incorrect images, leading to inconsistencies with the retrieved context. In Figure~\ref{fig:err_case2}, the model correctly retrieved and utilized relevant textual content in the answer, but incorrectly cited an unrelated table visual element. This issue stems from the inherent complexity of integrating visual and textual information, particularly when dealing with ambiguous or highly detailed visuals.

\clearpage
\section{M-GroSE performance by Context token length}
In this work, we evaluate seven models on our VinQA test set using two encoding strategies—\textit{Page Encoding} and \textit{Modality Encoding}—across five context‐token‐length intervals (0–2.5k, 2.5–5k, 5–7.5k, 7.5–10k, 10k–). Table \ref{tab:combined_with_f1} presents the overall M-GroSE performance across all models by context token length.

\section{Visual Source F1 performance by Modality type}
We evaluate seven models on our VinQA test set using two encoding strategies—\textit{Page Encoding} and \textit{Modality Encoding}—across four modality types (Table, Chart, Figure, Mixed). Table \ref{tab:single_modal_scores} shows the Visual Source performance of all models by modality type.

\begin{table*}[ht]
\centering
\tiny
\begin{tabular*}{\linewidth}{@{\extracolsep{\fill}} l | l | c c c c | c }
\toprule
\multirow{2}{*}{\textbf{Model}} 
  & \multirow{2}{*}{\textbf{Context Token Length}} 
  & \multicolumn{4}{c|}{\textbf{M-GroSE}} 
  & \multirow{2}{*}{\textbf{Avg}} \\
  & & \textbf{Relevancy} & \textbf{Completeness} & \textbf{Faithfulness} & \textbf{Unans. (F1)} & \\
\midrule

\rowcolor{gray!15}\multicolumn{7}{c}{\textit{\textbf{Page Encoding}}} \\
\midrule
\multirow{5}{*}{GPT-4.1}
  & 0–2500 & 4.88 & 3.41 & 4.40 & 0.72 & 3.42 \\
  & 2501–5000 & 4.91 & 3.80 & 4.41 & 0.79 & 3.53 \\
  & 5001–7500 & 4.95 & 3.92 & 4.44 & 0.81 & 3.58 \\
  & 7501–10000 & 4.85 & 3.74 & 4.23 & 0.80 & 3.46 \\
  & 10001– & 4.73 & 3.22 & 3.64 & 0.63 & 3.15 \\
\midrule
\multirow{5}{*}{GPT-4.1-mini}
  & 0–2500 & 4.69 & 3.38 & 4.21 & 0.57 & 3.32 \\
  & 2501–5000 & 4.66 & 3.62 & 4.07 & 0.49 & 3.34 \\
  & 5001–7500 & 4.61 & 3.83 & 4.14 & 0.45 & 3.4 \\
  & 7501–10000 & 4.45 & 3.61 & 3.91 & 0.50 & 3.25 \\
  & 10001– & 4.35 & 3.20 & 3.54 & 0.31 & 3.02 \\
\midrule
\multirow{5}{*}{Gemini 2.0 Flash}
  & 0–2500 & 4.42 & 2.95 & 4.30 & 0.90 & 3.17 \\
  & 2501–5000 & 4.09 & 2.78 & 3.85 & 0.85 & 2.93 \\
  & 5001–7500 & 4.33 & 2.65 & 3.77 & 0.90 & 2.94 \\
  & 7501–10000 & 4.31 & 2.57 & 3.47 & 0.86 & 2.84 \\
  & 10001– & 3.77 & 1.89 & 2.52 & 0.72 & 2.3 \\
\midrule
\multirow{5}{*}{Claude 3.5 Sonnet}
  & 0–2500 & 4.84 & 3.92 & 4.78 & 0.65 & 3.64 \\
  & 2501–5000 & 4.85 & 3.88 & 4.59 & 0.72 & 3.58 \\
  & 5001–7500 & 4.88 & 3.89 & 4.55 & 0.69 & 3.58 \\
  & 7501–10000 & 4.81 & 3.84 & 4.31 & 0.77 & 3.49 \\
  & 10001– & 4.71 & 3.39 & 3.97 & 0.59 & 3.27 \\
\midrule
\multirow{5}{*}{InternVL3-8B}
  & 0–2500 & 4.05 & 1.96 & 3.01 & 0.79 & 2.51 \\
  & 2501–5000 & 3.35 & 1.74 & 2.35 & 0.82 & 2.11 \\
  & 5001–7500 & 3.32 & 1.71 & 2.14 & 0.81 & 2.04 \\
  & 7501–10000 & 2.98 & 1.58 & 2.14 & 0.78 & 1.93 \\
  & 10001– & 2.63 & 1.25 & 1.61 & 0.53 & 1.62 \\
\midrule
\multirow{5}{*}{Qwen2.5-VL-7B}
  & 0–2500 & 3.42 & 1.76 & 2.89 & 0.74 & 2.27 \\
  & 2501–5000 & 2.79 & 1.61 & 2.41 & 0.67 & 1.95 \\
  & 5001–7500 & 3.18 & 1.68 & 2.31 & 0.70 & 2.04 \\
  & 7501–10000 & 3.03 & 1.62 & 2.25 & 0.76 & 1.97 \\
  & 10001– & 2.68 & 1.34 & 1.74 & 0.70 & 1.69 \\
\midrule
\multirow{5}{*}{\textbf{Qwen2.5-VL-7B (VinQA)}}
  & 0–2500 & 4.59 & 3.69 & 4.29 & 0.87 & 3.4 \\
  & 2501–5000 & 4.71 & 3.95 & 4.35 & 0.91 & 3.5 \\
  & 5001–7500 & 4.71 & 3.75 & 4.24 & 0.96 & 3.43 \\
  & 7501–10000 & 4.44 & 3.55 & 3.87 & 0.91 & 3.22 \\
  & 10001– & 4.40 & 3.03 & 3.47 & 0.81 & 2.98 \\
\midrule
\rowcolor{gray!15}\multicolumn{7}{c}{\textit{\textbf{Modality Encoding}}} \\
\midrule
\multirow{5}{*}{GPT-4.1}
  & 0–2500 & 4.92 & 3.84 & 4.54 & 0.67 & 3.58 \\
  & 2501–5000 & 4.93 & 4.05 & 4.62 & 0.68 & 3.65 \\
  & 5001–7500 & 4.89 & 4.05 & 4.64 & 0.70 & 3.65 \\
  & 7501–10000 & 4.84 & 4.03 & 4.60 & 0.76 & 3.62 \\
  & 10001– & 4.85 & 3.82 & 4.56 & 0.68 & 3.56 \\
\midrule
\multirow{5}{*}{GPT-4.1-mini}
  & 0–2500 & 4.74 & 3.72 & 4.34 & 0.75 & 3.45 \\
  & 2501–5000 & 4.63 & 3.98 & 4.33 & 0.77 & 3.49 \\
  & 5001–7500 & 4.69 & 4.00 & 4.22 & 0.77 & 3.48 \\
  & 7501–10000 & 4.55 & 4.08 & 4.06 & 0.77 & 3.42 \\
  & 10001– & 4.51 & 3.78 & 4.11 & 0.72 & 3.35 \\
\midrule
\multirow{5}{*}{Gemini 2.0 Flash}
  & 0–2500 & 4.65 & 3.41 & 4.58 & 0.89 & 3.41 \\
  & 2501–5000 & 4.62 & 3.50 & 4.54 & 0.89 & 3.42 \\
  & 5001–7500 & 4.80 & 3.62 & 4.56 & 0.85 & 3.5 \\
  & 7501–10000 & 4.67 & 3.51 & 4.56 & 0.89 & 3.44 \\
  & 10001– & 4.43 & 3.45 & 4.30 & 0.73 & 3.29 \\
\midrule
\multirow{5}{*}{Claude 3.5 Sonnet}
  & 0–2500 & 4.85 & 4.08 & 4.91 & 0.65 & 3.71 \\
  & 2501–5000 & 4.88 & 3.97 & 4.81 & 0.67 & 3.67 \\
  & 5001–7500 & 4.84 & 3.93 & 4.59 & 0.73 & 3.59 \\
  & 7501–10000 & 4.78 & 3.91 & 4.48 & 0.72 & 3.54 \\
  & 10001– & 4.79 & 3.81 & 4.62 & 0.61 & 3.56 \\
\midrule
\multirow{5}{*}{InternVL3-8B}
  & 0–2500 & 3.94 & 2.06 & 3.07 & 0.70 & 2.52 \\
  & 2501–5000 & 3.61 & 2.08 & 2.74 & 0.80 & 2.36 \\
  & 5001–7500 & 3.57 & 1.84 & 2.46 & 0.85 & 2.22 \\
  & 7501–10000 & 3.32 & 1.82 & 2.41 & 0.79 & 2.14 \\
  & 10001– & 2.86 & 1.57 & 2.17 & 0.54 & 1.90 \\
\midrule
\multirow{5}{*}{Qwen2.5-VL-7B}
  & 0–2500 & 3.22 & 1.79 & 2.66 & 0.64 & 2.17 \\
  & 2501–5000 & 2.72 & 1.66 & 2.32 & 0.75 & 1.93 \\
  & 5001–7500 & 3.01 & 1.85 & 2.53 & 0.79 & 2.10 \\
  & 7501–10000 & 3.03 & 1.87 & 2.60 & 0.81 & 2.12 \\
  & 10001– & 2.95 & 1.96 & 2.72 & 0.65 & 2.16 \\
\midrule
\multirow{5}{*}{\textbf{Qwen2.5-VL-7B (VinQA)}}
  & 0–2500 & 4.65 & 3.67 & 4.34 & 0.90 & 3.42 \\
  & 2501–5000 & 4.70 & 3.86 & 4.29 & 0.93 & 3.46 \\
  & 5001–7500 & 4.65 & 3.74 & 4.24 & 0.95 & 3.41 \\
  & 7501–10000 & 4.46 & 3.30 & 3.84 & 0.89 & 3.15 \\
  & 10001– & 4.40 & 3.14 & 3.57 & 0.80 & 3.03 \\
\bottomrule
\end{tabular*}
\caption{Overall M-GroSE performance across context token length.}
\label{tab:combined_with_f1}
\end{table*}


\begin{table*}[ht]
\centering
\tiny
\begin{tabular*}{\linewidth}{@{\extracolsep{\fill}} l | l | c c c}
\toprule
\multirow{2}{*}{\textbf{Model}} & \multirow{2}{*}{\textbf{Modal Type}} & \multicolumn{3}{c}{\textbf{Visual Source}} \\
 & & \textbf{Precision} & \textbf{Recall} & \textbf{F1} \\
\midrule
\rowcolor{gray!15}
\multicolumn{5}{c}{\textit{\textbf{Page Encoding}}} \\
\midrule
\multirow{4}{*}{GPT-4.1}
& Table  & 72.86 & 54.51 & 62.37 \\
& Chart  & 70.12 & 58.81 & 63.97 \\
& Figure & 71.92 & 51.47 & 60.00 \\
& Mixed  & 83.25 & 46.17 & 59.40 \\
\midrule
\multirow{4}{*}{GPT-4.1-mini}
& Table  & 62.94 & 37.03 & 46.63 \\
& Chart  & 68.21 & 54.84 & 60.80 \\
& Figure & 66.83 & 25.96 & 37.40 \\
& Mixed  & 85.71 & 36.07 & 50.77 \\
\midrule
\multirow{4}{*}{Gemini 2.0 Flash}
& Table  & 65.93 & 33.46 & 44.39 \\
& Chart  & 68.18 & 40.94 & 51.16 \\
& Figure & 72.63 & 31.15 & 43.60 \\
& Mixed  & 83.33 & 32.79 & 47.06 \\
\midrule
\multirow{4}{*}{Claude 3.5 Sonnet}
& Table  & 72.87 & 53.01 & 61.37 \\
& Chart  & 70.03 & 62.03 & 65.79 \\
& Figure & 70.63 & 62.98 & 66.59 \\
& Mixed  & 83.69 & 53.28 & 65.11 \\
\midrule
\multirow{4}{*}{InternVL3-8B}
& Table  & 50.72 & 6.58 & 11.65 \\
& Chart  & 70.29 & 24.07 & 35.86 \\
& Figure & 71.25 & 12.87 & 21.80 \\
& Mixed  & 84.62 & 15.03 & 25.52 \\
\midrule
\multirow{4}{*}{Qwen2.5-VL-7B}
& Table  & 51.61 & 15.04 & 23.29 \\
& Chart  & 70.83 & 33.75 & 45.71 \\
& Figure & 72.15 & 12.87  & 21.84 \\
& Mixed  & 77.66 & 19.95 & 31.74 \\
\midrule
\multirow{4}{*}{\textbf{Qwen2.5-VL-7B (VinQA)}}
& Table  & 77.29 & 39.66 & 52.42 \\
& Chart  & 72.36 & 49.38 & 58.70 \\
& Figure & 74.60 & 41.76 & 53.55 \\
& Mixed  & 85.80 & 41.26 & 55.72 \\
\midrule
\rowcolor{gray!15}
\multicolumn{5}{c}{\textit{\textbf{Modality Encoding}}} \\
\midrule
\multirow{4}{*}{GPT-4.1}
& Table  & 84.30 & 65.60 & 73.78 \\
& Chart  & 70.78 & 65.51 & 68.04 \\
& Figure & 72.83 & 72.01 & 72.42 \\
& Mixed  & 83.65 & 63.93 & 73.49 \\
\midrule
\multirow{4}{*}{GPT-4.1-mini}
& Table  & 77.26 & 56.20 & 65.07 \\
& Chart  & 71.75 & 64.27 & 67.80 \\
& Figure & 65.95 & 34.54 & 45.33 \\
& Mixed  & 85.71 & 45.90 & 59.79 \\
\midrule
\multirow{4}{*}{Gemini 2.0 Flash}
& Table  & 84.49 & 57.33 & 68.31 \\
& Chart  & 69.97 & 52.61 & 60.06 \\
& Figure & 75.10 & 42.21 & 54.05 \\
& Mixed  & 88.44 & 41.80 & 56.77 \\
\midrule
\multirow{4}{*}{Claude 3.5 Sonnet}
& Table  & 85.10 & 63.25 & 72.63 \\
& Chart  & 72.83 & 66.50 & 69.52 \\
& Figure & 70.24 & 65.01 & 67.53 \\
& Mixed  & 85.66 & 62.02 & 71.95 \\
\midrule
\multirow{4}{*}{InternVL3-8B}
& Table  & 75.83 & 17.11 & 31.51 \\
& Chart  & 76.05 & 31.51 & 44.56 \\
& Figure & 68.04 & 14.90 & 24.44 \\
& Mixed  & 88.31 & 18.58 & 30.70 \\
\midrule
\multirow{4}{*}{Qwen2.5-VL-7B}
& Table  & 84.80 & 27.26 & 41.25 \\
& Chart  & 71.90 & 37.47 & 49.27 \\
& Figure & 63.30 & 15.58 & 25.00 \\
& Mixed  & 83.61 & 27.87 & 41.80 \\
\midrule
\multirow{4}{*}{\textbf{Qwen2.5-VL-7B (VinQA)}}
& Table  & 82.94 & 45.68 & 58.91 \\
& Chart  & 71.94 & 55.33 & 62.55 \\
& Figure & 71.03 & 40.41 & 51.51 \\
& Mixed  & 83.16 & 43.17 & 56.83 \\
\bottomrule
\end{tabular*}
\caption{Overall Visual Source performance across modality types.}
\label{tab:single_modal_scores}
\end{table*}

\end{document}